\documentclass{agujournal2019}
\usepackage{url}
\usepackage{lineno}
\usepackage[inline]{trackchanges}
\usepackage{soul}
\usepackage{booktabs}
\usepackage{multirow}
\usepackage{amsmath}
\usepackage{microtype}
\usepackage{mleftright}
\mleftright
\journalname{Journal of Advances in Modeling Earth Systems (JAMES)}

\begin{document}
\RaggedRight



\title{Discovering Equations for the Thermodynamics of South Asian Monsoon Precipitation}

\authors{Savannah L. Ferretti\affil{1}, Jane W. Baldwin\affil{1}, Michael S. Pritchard\affil{2}, and Tom Beucler\affil{3,4}}

\affiliation{1}{Department of Earth System Science, University of California, Irvine, Irvine, CA, United States}
\affiliation{2}{NVIDIA Corporation, Santa Clara, CA, United States}
\affiliation{3}{Faculty of Geosciences and Environment, University of Lausanne, Lausanne, VD, Switzerland}
\affiliation{4}{Expertise Centre for Climate Extremes, University of Lausanne, Lausanne, VD, Switzerland}

\correspondingauthor{Savannah L. Ferretti}{savannah.ferretti@asu.edu}

\begin{keypoints}
    \item Integration kernel learning with symbolic regression yields physically consistent equations for South Asian monsoon precipitation.
    \item Precipitation reflects competition between moisture and instability, with moisture distinguishing the heaviest precipitation states.
    \item Coupling sensible heat flux to near-surface thermodynamics improves predictions of ocean precipitation and its diurnal variability.
\end{keypoints}


\begin{abstract}

South Asian monsoon precipitation depends on the thermodynamic structure of the lower troposphere, but how its ingredients combine to shape precipitation remains unclear, partly because this dependence is nonlocal in the vertical. We use symbolic distillation, which combines neural networks and symbolic regression, to discover interpretable precipitation equations from 21 years of June--August reanalysis data. Neural networks first learn the vertical thermodynamic structure most relevant to precipitation, condensing each variable into a single interpretable feature not tied to a particular vertical grid. Then, symbolic regression derives closed-form equations from these features and the surface state. Our best equation explains 1.5 times as much precipitation variance as a widely used buoyancy-based diagnostic and recovers about 85\% of neural network skill while using about a thousand times fewer parameters. Unlike the buoyancy diagnostic, which adds moisture and instability, the equations treat them as competing controls, with the heaviest precipitation occurring when the lower free troposphere is humid. Over ocean, sensible heating is associated with additional precipitation mainly when near-surface air is warm and moist, improving predicted precipitation and its diurnal cycle. Although the learned features were optimized only for precipitation prediction, the equations largely respond to each feature as expected from convective physics, consistent with the reanalysis. These responses reduce the dependence of monsoon precipitation on the entire vertical column to a few simple, testable constraints that could guide data-driven precipitation models to make them more physically consistent.

\end{abstract}

\section*{Plain Language Summary}

Monsoon rain over South Asia depends on how warm and moist the lower atmosphere is, and how easily air can rise, but scientists still debate how these factors work together. Using two decades of summer weather data, we trained artificial intelligence (AI) models to learn which layers of the atmosphere matter most for rain, and then searched for short, easy-to-understand equations that describe how rain depends on them. These equations retain much of the skill of the far more complex AI models. They show that moist air and the tendency of air to rise act as competing ingredients, so rain depends mainly on whichever is stronger. The heaviest rain occurs when the air a few kilometers above the surface is very humid. Over ocean, heat rising from the sea surface coincides with more rain, but mostly when the air near the surface is already warm and moist. The equations also mostly follow the basic physics of storms, even though this behavior was never built into them. These simple rules could help make other precipitation models more realistic.

\section{Introduction}\label{sec:introduction}

South Asian monsoon precipitation depends on the thermodynamic structure of the atmosphere, but it remains unclear how that dependence operates. One line of work emphasizes lower free-tropospheric moisture regulation, in which precipitation onset, progression, and withdrawal are governed by the interaction between moist low-level flow and overlying dry air \cite{Parker2016, Volonte2020, Menon2022, Deoras2024}. A second perspective emphasizes boundary layer energy and instability, particularly for extreme precipitation over central India, where moisture preconditioning and convective potential play important roles in regulating precipitation \cite{Bajrang2023, Masiwal2023}. At the same time, precipitation exhibits departures from simple thermodynamic scaling, with both sub- and super-Clausius--Clapeyron behavior reported depending on region and timescale \cite{Ali2017, Ghausi2022, Varghese2024, Andria2025}. These results indicate that while the relevant thermodynamic ingredients are known, there is no consensus on how they combine to shape precipitation across the domain.

A widely used process-oriented diagnostic for lower-tropospheric buoyancy, $B_L$ \cite{Ahmed2020, Ahmed2021}, is defined as the scaled and weighted difference between the instability of boundary layer air relative to the lower free troposphere and the dryness of the lower free troposphere. For South Asian monsoon precipitation, $B_L$ has been used to characterize spatiotemporal variability over the Western Ghats \cite{Nicolas2024}, explain regional monsoon seasonality \cite{Ferretti2025}, and link El Ni\~{n}o to intensified extreme daily precipitation through changes in convective buoyancy \cite{Hill2025}. Its appeal lies in reducing lower-tropospheric thermodynamic structure to a single physically interpretable quantity. However, this simplification rests on several assumptions:
\begin{itemize}
    \item Precipitation variability depends mainly on these two thermodynamic properties.
    \item These properties can be evaluated from averages over a boundary layer and a lower free troposphere separated by a prescribed boundary.
    \item A prescribed entrainment profile determines how much each layer contributes to convection.
    \item Buoyancy can be estimated from a scaled difference in equivalent potential temperature, even though the scaling changes with temperature.
\end{itemize}
These assumptions may limit the diagnostic's ability to capture regional differences in thermodynamic structure and convective behavior.

The goal of this study is to revisit the assumptions underlying $B_L$ by using data-driven techniques to discover alternative thermodynamic expressions for South Asian monsoon precipitation, and to identify how much predictive skill can be retained in simple, interpretable equations. Symbolic regression searches directly for closed-form relationships that balance predictive skill and complexity \cite{Cranmer2023}, and has recovered physically meaningful relationships and reduced-order parameterizations across a number of Earth system applications \cite{Hafner2023,Grundner2024,Tam2025,Xu2025}. However, symbolic regression scales poorly with the number of inputs, because the space of candidate equations grows combinatorially with each one added. This limitation is acute for precipitation, which depends nonlocally on the vertical thermodynamic structure, since treating each variable at each vertical level as a separate input creates a search space far too large to explore. We therefore adapt symbolic distillation \cite{Cranmer2020}, in which a neural network first learns a compact representation of high-dimensional inputs and symbolic regression then derives closed-form equations from that representation. First, integration kernel learning \cite{Ferretti2026} trains a neural network to predict precipitation while learning how to weight each thermodynamic variable in the vertical, compressing each profile into a single scalar feature. This amounts to data-driven dimensionality reduction over the vertical column, guided by predictive skill rather than prescribed layers. Second, symbolic regression condenses these few learned features, along with surface variables, into closed-form equations fit directly to reanalysis precipitation rather than to the predictions of the neural network. This allows both the relevant vertical thermodynamic structure and its functional relationship with precipitation to be determined from the data rather than prescribed. Comparing the resulting equations with $B_L$ and more flexible neural network models helps assess the tradeoff between predictive skill and complexity \cite{Mamalakis2026} and clarify how monsoon precipitation depends on atmospheric and surface conditions across South Asia.

The remainder of this study is structured as follows. Section~\ref{sec:methods} introduces the data and methodology. Section~\ref{sec:results} evaluates performance across the model hierarchy and interprets the discovered equations in terms of the physical processes they represent. Section~\ref{sec:conclusion} summarizes the main findings and their implications.

\section{Methods}\label{sec:methods}

\subsection{Data}\label{subsec:data}

All predictors are derived from the ERA5 reanalysis \cite{Hersbach2020}, which accurately represents thermodynamic structure over the South Asian monsoon region \cite{Johnston2021}. The predictor set includes both surface variables and vertical profiles of thermodynamic quantities. For surface variables, we include land fraction ($\mathrm{LF}$), latent heat flux ($\mathrm{LHF}$), and sensible heat flux ($\mathrm{SHF}$; \citeauthor{Hersbach2023b}, \citeyearNP{Hersbach2023b}). We reverse the ERA5 flux sign convention during preprocessing so that positive $\mathrm{LHF}$ and $\mathrm{SHF}$ indicate upward energy transfer from the surface to the atmosphere, and retain them separately because their partitioning, commonly expressed by the Bowen ratio, distinguishes energy used for boundary layer moistening from energy used for thermal destabilization \cite{Bowen1926}. For vertical profiles, we use relative humidity ($\mathrm{RH}$), equivalent potential temperature ($\theta_e$), and saturated equivalent potential temperature ($\theta_e^*$), derived from air temperature and specific humidity on pressure levels between 1{,}000 and 500~hPa, together with surface pressure \cite{Hersbach2023a}. We use these variables rather than temperature and specific humidity directly because tropical convection depends more strongly on moist parcel energetics, environmental humidity, and saturation state than on temperature or moisture independently. Although $\mathrm{RH}$, $\theta_e$, and $\theta_e^*$ contain overlapping information, retaining all three allows the learning framework to determine which representation of moist thermodynamic structure yields the simplest and most predictive relationships with precipitation. We restrict the profiles to the 1{,}000--500~hPa layer for consistency with the $B_L$ formulation \cite{Ahmed2021} and because convective development is most sensitive to lower-tropospheric thermodynamic structure, whereas structure above 500~hPa is more strongly modified by convection itself. We also compute $B_L$ (m/s\textsuperscript{2}), the sole predictor in two models (Section~\ref{subsec:experiments}), following \citeA{Ahmed2021} and \citeA{Ferretti2025}:
\begin{equation}
    B_L = \frac{g}{\kappa_L\theta_{e0}}\left[w_B\left(\frac{\theta_{eB}-{\theta_{eL}}^{*}}{{\theta_{eL}}^{*}}\right)\theta_{e0}-w_L\left(\frac{{\theta_{eL}}^{*}-\theta_{eL}}{{\theta_{eL}}^{*}}\right)\theta_{e0}\right],
    \label{eq:bl}
\end{equation}
where subscripts $B$ and $L$ denote averages over the boundary layer (the lowest 100~hPa above the surface) and the lower free troposphere (from the top of the boundary layer to 500~hPa), $\kappa_L$ is a scaling constant, $\theta_{e0}$ is a reference $\theta_e$, and $w_B$ and $w_L$ are layer weights set by a prescribed entrainment profile (see \citeA{Ferretti2025} for more details).

To minimize the influence of below-ground pressure levels that arise from the substantial orography across the South Asian monsoon domain, we linearly interpolate $\mathrm{RH}$, $\theta_e$, and $\theta_e^*$ in pressure from the 1{,}000--500~hPa levels to 11 evenly spaced terrain-following sigma ($\sigma = p/p_s$) levels over $0.5 \leq \sigma \leq 1$. Where a sigma level lies above 500~hPa (over elevated terrain) or below 1{,}000~hPa (mostly over ocean), we use the value at the nearest pressure level instead of extrapolating. This transformation represents vertical thermodynamic structure relative to the local surface, allowing the learned vertical weightings (Section~\ref{subsubsec:kernels}) to more consistently identify comparable portions of the lower troposphere across regions of complex terrain. Similar terrain-following representations have also been shown to improve near-surface structure and precipitation processes in AI weather prediction models \cite{Sha2025}. Quadrature weights for vertical integration on the sigma grid ($\Delta\sigma$) are computed using centered finite differences.

ERA5 total precipitation is used as the prediction target. Although precipitation itself is not assimilated, it is generated from an atmospheric state strongly constrained by assimilated observations, so the predictors and target are physically consistent and defined on the same grid and time windows. Satellite precipitation estimates offer more direct observational constraints but are independent of the ERA5 thermodynamic state and carry their own retrieval uncertainties, so pairing them with ERA5 predictors would introduce mismatches between the target and the state from which the equations are derived. ERA5 precipitation is nonetheless produced by a convection parameterization and exhibits known regional and intensity-dependent biases over the South Asian monsoon region \cite{Hassler2021}, so the discovered equations describe precipitation as represented in ERA5.

We use data from the peak monsoon season (June--August) for 2000--2020 over the South Asian monsoon region (5\textdegree--25\textdegree N, 60\textdegree--90\textdegree E). ERA5 data are available hourly on a 0.25\textdegree\,\texttimes\,0.25\textdegree{} grid. We interpolate the data to 1.0\textdegree\,\texttimes\,1.0\textdegree{} and average over 3-hour windows. This spatiotemporal resolution emphasizes sustained and organized convective systems rather than individual convective cells, since it is at these larger scales that precipitation exhibits stronger deterministic relationships with thermodynamic structure and is typically parameterized in climate models. Predictors are computed from hourly data and averaged over each window, and precipitation is accumulated over the same window. The learned relationships are therefore concurrent and should not be interpreted as causal, since the window-averaged state also reflects the influence of convection, and surface fluxes can respond to precipitation (e.g., through cold pools). Precipitation below 10\textsuperscript{--4}~mm per window is set to zero to remove small numerical residuals. The resulting dataset contains 651 spatial grid points and 15{,}456 timesteps (approximately 10 million samples total). We split these data chronologically, training on 2000--2014, validating on 2015--2017, and testing on 2018--2020.

\subsection{Data-Driven Models}\label{subsec:models}

We construct data-driven models that differ in three modeling choices: (1) predictor set, (2) representation of vertical structure, and (3) functional form relating predictors to precipitation. Neural networks provide flexible mappings from predictors to precipitation, while integration kernel learning and symbolic regression form the two stages of symbolic distillation, compressing vertical structure into interpretable scalar features and then deriving closed-form equations for precipitation.

\subsubsection{Prediction Framework}\label{subsubsec:framework}

Because precipitation spans several orders of magnitude and predictors differ in units and scale, all models operate on standardized inputs and a transformed precipitation target. Each predictor is standardized to zero mean and unit variance using its training set mean ($\mu$) and standard deviation ($s$), except $\mathrm{LF}$, which is not standardized because it is time-invariant and bounded within $[0,1]$.

The precipitation target is transformed in two steps. First, we apply $y = \log(1+\mathrm{P})$, where $\mathrm{P}$ is precipitation in mm, to compress the heavy-tailed distribution. Second, we standardize $y$ to zero mean and unit variance using $\mu_y$ and $s_y$. Each model is then trained to minimize the mean squared error (MSE) between its standardized prediction and the standardized $y$. We use MSE because, in preliminary experiments, it yielded higher R\textsuperscript{2} than mean absolute error, Huber, Tweedie, and quantile losses. Because this error is minimized in log space, predictions approximate the conditional mean of $\log(1+\mathrm{P})$, which underestimates the conditional mean of $\mathrm{P}$ once transformed back, most strongly where precipitation is heavy and variable. This limitation is shared by all models, so it affects absolute precipitation magnitudes more than comparisons among models.

All models enforce non-negative predictions through the same mechanism. Each model produces a scalar output $z$, and its standardized prediction is the standardized value of zero precipitation plus the positive part of $z$, $z^+ = \max(z,0)$. Non-positive $z$ therefore predicts zero precipitation, and positive $z$ predicts precipitation that increases with $z$. Inverting the standardization and log-transformation (Text~S1 of Supporting Information) gives physical precipitation:
\begin{equation}
    \mathrm{P} = (\exp(s_yz)-1)^+.
    \label{eq:prediction}
\end{equation}

\subsubsection{Neural Networks}\label{subsubsec:nn}
 
All neural network models use the architecture and training configuration described in Appendix~B of \citeA{Ferretti2026}. Briefly, each is a fully connected feedforward network with four hidden layers of widths 256, 128, 64, and 32, GELU activations, and dropout with probability 0.1, trained using the Adam optimizer. Each model produces $z$ in Equation~\ref{eq:prediction}, with non-negativity enforced by construction. We train each model using three independent random seeds (42, 72, and 102).

\subsubsection{Integration Kernel Learning}\label{subsubsec:kernels}

Neural networks can ingest vertically resolved predictors directly, whereas symbolic regression is more effective with a small number of scalar inputs. We therefore use the integration kernel learning framework \cite{Ferretti2026} to compress each thermodynamic profile into a single interpretable feature. Integration kernels represent vertical structure through learned weighted integrals, allowing the model to identify which portions of each profile are most relevant to precipitation prediction.

Following \citeA{Ferretti2026}, we denote each vertically varying predictor ($\mathrm{RH}$, $\theta_e$, or $\theta_e^*$) by $\varphi_i(\sigma)$, where $i$ indexes the predictor and $\sigma$ is the vertical (sigma) level. For each predictor, the kernel-integrated feature $\widehat{\varphi}_i$ is the weighted integral of the profile over the vertical column:
\begin{equation}
    \widehat{\varphi}_i = \int_{0.5}^1k_i(\sigma)\varphi_i(\sigma)\mathrm{d}\sigma,
    \label{eq:kernelcontinuous}
\end{equation}
where $k_i(\sigma)$ is a learned kernel weighting function that sets how much each level contributes. When $k_i(\sigma)$ is defined as a continuous function, the resulting equations can be evaluated on any vertical grid without changing their mathematical form. On the discrete sigma grid used in practice, Equation~\ref{eq:kernelcontinuous} becomes:
\begin{equation}
    \widehat{\varphi}_i = \sum_{m=1}^{N_\sigma}k_i(\sigma_m)\varphi_i(\sigma_m)\Delta\sigma_m,
    \label{eq:kerneldiscrete}
\end{equation}
where $k_i(\sigma_m)$ is the learned kernel weight at sigma level $\sigma_m$, $N_\sigma$ is the number of sigma levels, and $\Delta\sigma_m$ is the vertical quadrature weight. Each kernel is normalized to integrate to unity over the vertical column, so $\widehat{\varphi}_i$ is a kernel-weighted vertical integral that retains the physical units of $\varphi_i$. We denote the resulting features for $\mathrm{RH}$, $\theta_e$, and $\theta_e^*$ by $\widehat{\mathrm{RH}}$, $\widehat{\theta_e}$, and $\widehat{\theta_e^*}$. The kernels are learned jointly with the downstream neural network, which predicts precipitation from each feature standardized as described in Section~\ref{subsubsec:framework}. Because each variable is standardized using a single mean and standard deviation pooled over all sigma levels, grid points, and timesteps in the training set, and each kernel integrates to unity, standardizing before or after kernel integration yields the same feature.

We consider both nonparametric and Gaussian kernel parameterizations. For nonparametric kernels, $k_i(\sigma_m)$ is learned independently at each sigma level, allowing flexible weighting across the vertical profile. For Gaussian kernels, $k_i(\sigma_m)$ is obtained by evaluating a continuous Gaussian function with learned center and width parameters at each sigma level, constraining each predictor to a single dominant vertical region. The Gaussian parameterization is used for symbolic regression because its simpler vertical structure is more physically interpretable. Additional details on the Gaussian parameterization, including its definition, learned kernel shapes, and its selection over the nonparametric alternative, are provided in Text~S2, Table~S1, and Figure~S1 of Supporting Information.

\subsubsection{Symbolic Regression}\label{subsubsec:sr}

We use the PySR package \cite{Cranmer2023} to derive closed-form equations for $z$ in Equation~\ref{eq:prediction}. Rather than assuming a functional form \textit{a priori}, PySR uses genetic programming to search for compact equations that balance predictive skill against complexity. We use it in two ways. In the first, it discovers a complete equation for $z$ from a selected set of predictors. In the second, it receives the output of an existing equation as an additional input, so that it can search for a new equation that incorporates or extends the existing one. The predictor sets for each search are described in Section~\ref{subsec:experiments}.

Because symbolic regression scales poorly with sample count, each search uses a stratified subset of the training and validation data and is repeated with three random seeds (42, 72, and 102). Each seed produces a Pareto frontier, the set of candidate equations whose skill cannot improve without increasing complexity. We select representative equation structures based on predictive skill, physical interpretability, and, where possible, consistency across seeds, since a structure that recurs independently across seeds is more likely to reflect a robust feature of the underlying relationship than an artifact of the search.

We then optimize the constants of each selected equation on the combined training and validation sets. For searches that build on an existing equation, the constants of the existing equation are optimized first and then held fixed, so that only the new terms are fitted. Because PySR operates on standardized predictors and the transformed target, the resulting equations are in standardized form. For interpretation, we rewrite each one in physical units by substituting the standardization of each predictor and combining constants into physically meaningful parameters. All equations in Section~\ref{sec:results} are presented in this physical form. The search configuration, complexity measure, subsampling strategy, and constant optimization are detailed in Text~S3 of Supporting Information, and the physical constants of each model, including their values and units, are given in Table~S2.

\subsection{Experimental Design}\label{subsec:experiments}

We evaluate nine models spanning neural network and symbolic regression approaches to isolate the roles of predictor choice, vertical structure, and functional form.

We first evaluate four neural network models that share the architecture described in Section~\ref{subsubsec:nn} but differ in their inputs. NN-BL uses $B_L$ alone. NN-FULL uses the vertically resolved thermodynamic predictors $\mathrm{RH}$, $\theta_e$, and $\theta_e^*$ together with the surface predictors $\mathrm{LF}$, $\mathrm{LHF}$, and $\mathrm{SHF}$. NN-NONPARAM and NN-GAUSS retain the surface predictors but replace each vertically resolved predictor with a scalar feature obtained by integration with its learned nonparametric or Gaussian kernel. NN-BL thus tests how much predictive information the buoyancy diagnostic retains, NN-FULL provides a reference for the skill achievable from the full predictor set, and NN-NONPARAM and NN-GAUSS test how much of that skill is retained when each vertically varying predictor is compressed into a kernel-integrated feature.

We then evaluate four symbolic regression models obtained with the procedure of Section~\ref{subsubsec:sr}. SR-BL serves as a baseline that applies symbolic regression directly to a prescribed feature, learning a closed-form relationship between $B_L$ and precipitation. The remaining models instead distill features learned by NN-GAUSS. SR-ATM discovers a complete equation from the Gaussian-kernel-integrated features $\widehat{\mathrm{RH}}$, $\widehat{\theta_e}$, and $\widehat{\theta_e^*}$ to identify an interpretable relationship based only on the thermodynamic state. SR-SFC and SR-ALL build on SR-ATM, receiving its output as an input and adding a learned correction. SR-SFC's correction draws only on $\mathrm{LF}$, $\mathrm{LHF}$, and $\mathrm{SHF}$, testing whether surface conditions account for the errors left by SR-ATM. SR-ALL's correction has access to the full predictor set, including both the kernel-integrated features and the surface variables. 

Finally, we construct a ninth model, SR-ALL-PC, by structurally modifying the form of SR-ALL to improve its physical consistency and then reoptimizing its constants. Unlike the other symbolic regression models, its form is not discovered directly by PySR. The motivation for this modification and its construction are described in Section~\ref{subsec:srall}.

\section{Results}\label{sec:results}

\subsection{Model Skill and the Complexity--Performance Tradeoff}\label{subsec:pareto}

\begin{figure}[h!] 
    \centering
    \includegraphics[width=\textwidth]{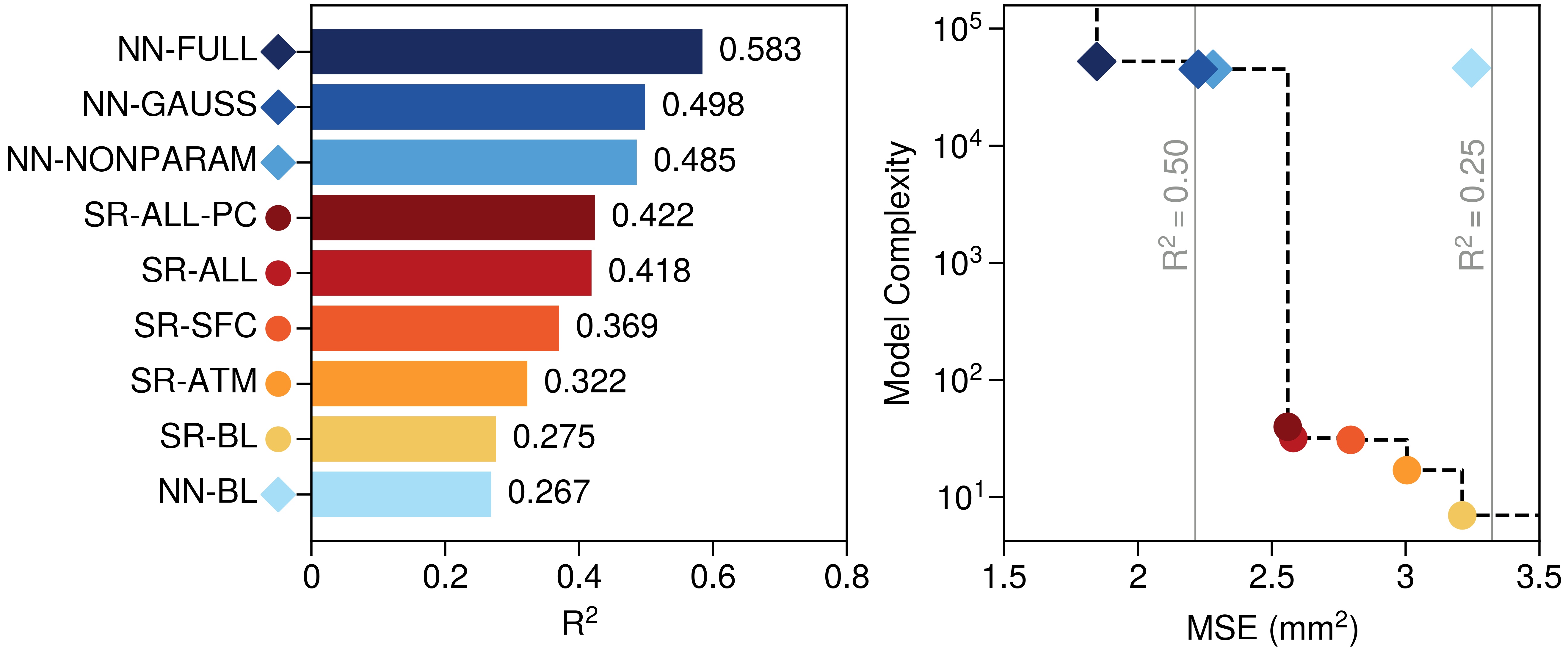} 
    \caption{(Left) Test set R\textsuperscript{2} for each model in the hierarchy. (Right) Test set MSE versus model complexity. The black dashed line traces the Pareto frontier across all models, and gray vertical lines indicate constant-R\textsuperscript{2} levels. For neural networks, R\textsuperscript{2} and MSE are averaged across seeds (\textit{n} = 3), with the extremely small inter-seed variability omitted. Model complexity is shown on a log scale, which compresses differences among neural network models that have parameter counts within a factor of two of each other.}
    \label{fig:1}
\end{figure}

Figure~\ref{fig:1} summarizes model performance across the hierarchy, showing test set R\textsuperscript{2} (left) and the complexity--performance tradeoff (right). R\textsuperscript{2} measures the fraction of precipitation variance explained, and MSE measures the magnitude of prediction error in physical precipitation (mm\textsuperscript{2}). To facilitate comparison between neural networks and symbolic equations, complexity is measured by trainable parameter count for neural networks, including nonparametric kernel weights and Gaussian shape parameters, and by weighted node count for symbolic equations, the measure symbolic regression uses to balance skill against complexity during the search (Text~S3 of Supporting Information). The complexity of SR-SFC, SR-ALL, and SR-ALL-PC includes that of SR-ATM, on which they build.

The full neural network (NN-FULL) achieves the highest skill (R\textsuperscript{2} $\approx$ 0.58), providing a reference for the skill achievable from the full predictor set with this architecture. Restricting the neural network input to $B_L$ alone (NN-BL) substantially reduces performance (R\textsuperscript{2} $\approx$ 0.27), consistent with the hypothesis that $B_L$ does not retain all the predictive information in the full thermodynamic structure, though part of this reduction may also arise from the exclusion of surface predictors present in NN-FULL.

Neural networks trained on kernel-integrated features and local surface predictors recover much of NN-FULL's predictive skill at lower complexity. NN-NONPARAM and NN-GAUSS achieve R\textsuperscript{2} $\approx$ 0.49 and 0.50, respectively, using approximately 45{,}000 trainable parameters compared with approximately 53{,}000 for NN-FULL. Their close agreement in skill is discussed further in Text~S2 of Supporting Information. These results suggest that a small number of \textit{learned} vertically integrated features retain much of the predictive information in the full thermodynamic structure, although these networks remain difficult to interpret directly because they require tens of thousands of parameters.

The symbolic regression models test how much predictive skill can be retained in interpretable analytic expressions. With $B_L$ as input, SR-BL performs comparably to NN-BL (R\textsuperscript{2} $\approx$ 0.28 versus 0.27), indicating that a compact closed-form expression is sufficient to capture the relationships available from this predictor. Replacing $B_L$ with the kernel-integrated features raises skill to R\textsuperscript{2} $\approx$ 0.32 in SR-ATM, with a larger gain over land (from 0.21 to 0.29) than over ocean (from 0.32 to 0.34), and because neither model uses surface predictors, this gain confirms that $B_L$ does not retain all of the predictive information in the lower-tropospheric thermodynamic structure. Adding surface predictors (SR-SFC and SR-ALL) further raises skill (R\textsuperscript{2} $\approx$ 0.37 and 0.42, respectively), allowing relatively simple equations to recover a substantial fraction of NN-GAUSS's predictive benefit with far fewer parameters. This skill is preserved when SR-ALL is structurally modified to improve its physical consistency (Section~\ref{subsec:srall}), with SR-ALL-PC achieving R\textsuperscript{2} $\approx$ 0.42.

The Pareto frontier reveals a relatively sharp transition between the symbolic regression models and the neural network models, with complexity increasing by roughly three orders of magnitude between SR-ALL-PC and NN-GAUSS for a comparatively modest gain in skill. This suggests that relatively simple thermodynamic relationships capture a substantial fraction of South Asian monsoon precipitation variability. Therefore, our focus remains on the more interpretable symbolic expressions rather than the more complex neural networks.

\subsection{Buoyancy as a Baseline Thermodynamic Predictor}\label{subsec:srbl}

SR-BL provides a baseline in which both the relevant vertical layers and the way they combine are prescribed by $B_L$. \citeA{Ahmed2020} approximated the relationship between precipitation and $B_L$ with a piecewise-linear ramp, in which precipitation is zero below a critical threshold and increases linearly above it. SR-BL instead learns this relationship from the data, taking the form:
\begin{equation}
    \mathrm{P} = 
    \left(\exp\left(\lambda\left(B_L-B_c\right)^3+\beta\right)-1\right)^+,
    \label{eq:srbl}
\end{equation}
where $B_c$ is a critical buoyancy threshold below which precipitation is negligible, $\lambda$ scales the cubed departure of $B_L$ from $B_c$, and $\beta$ sets the precipitation at the critical threshold (Table~S2 of Supporting Information).

\begin{figure}[h!]
    \centering
    \includegraphics[width=2.48in]{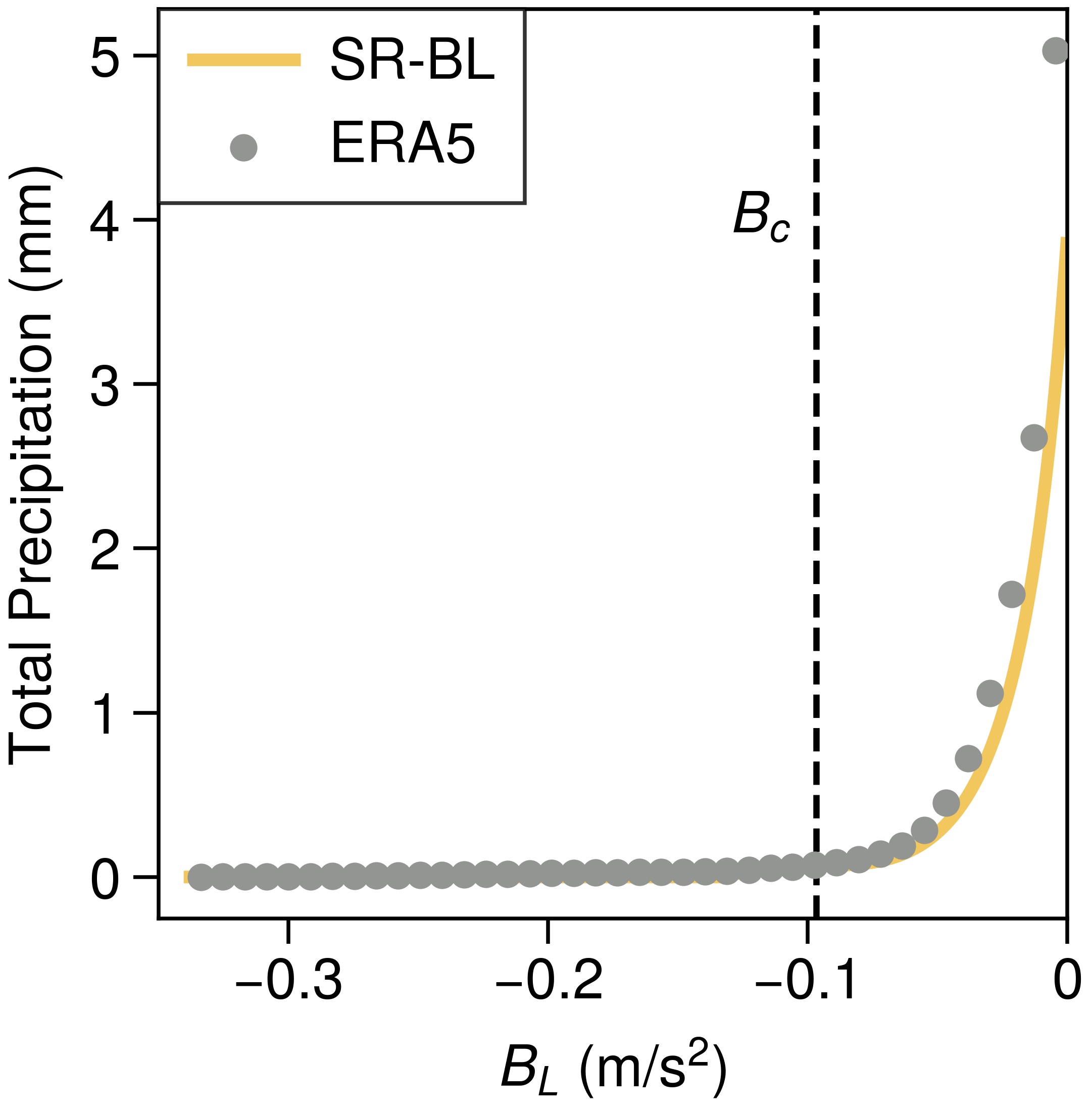}
    \caption{Precipitation response as a function of $B_L$ in the test set. Gray circles show bin-mean ERA5 3-hourly total precipitation (mm) for bins containing at least 50 samples. The yellow curve shows SR-BL (Equation~\ref{eq:srbl}), and the black dashed line marks $B_c$.}
    \label{fig:2}
\end{figure}

Figure~\ref{fig:2} shows the precipitation response learned by SR-BL on the test set, together with bin-mean ERA5 precipitation as a function of $B_L$. The transition from suppressed to active convection is smooth and accelerating, consistent with conceptual frameworks in which buoyancy controls convective intensity \cite{Ahmed2020, Ahmed2021, Schiro2018}. However, unlike the linear ramp, SR-BL discovers a cubic-exponential response that produces progressively amplifying precipitation increases as buoyancy becomes more favorable.

Despite capturing this convective transition, SR-BL explains only about a quarter of the test set precipitation variance (Figure~\ref{fig:1}). Its underestimation of the most extreme bins partly reflects training in log space (Section~\ref{subsubsec:framework}), but its limited overall skill suggests that a single buoyancy measure does not retain all of the thermodynamic information relevant to monsoon precipitation, motivating us to learn the relevant vertical structure directly from the data, which we examine next.

\subsection{Learned Vertical Thermodynamic Structure}\label{subsec:structure}

The limitations of SR-BL raise the question of which portions of the lower troposphere precipitation actually depends on, and whether they coincide with the layers prescribed in $B_L$. Integration kernel learning addresses this by learning the vertical weighting of each thermodynamic variable directly from the data \cite{Ferretti2026}.

\begin{figure}[h!]
    \centering
    \includegraphics[width=5.13in]{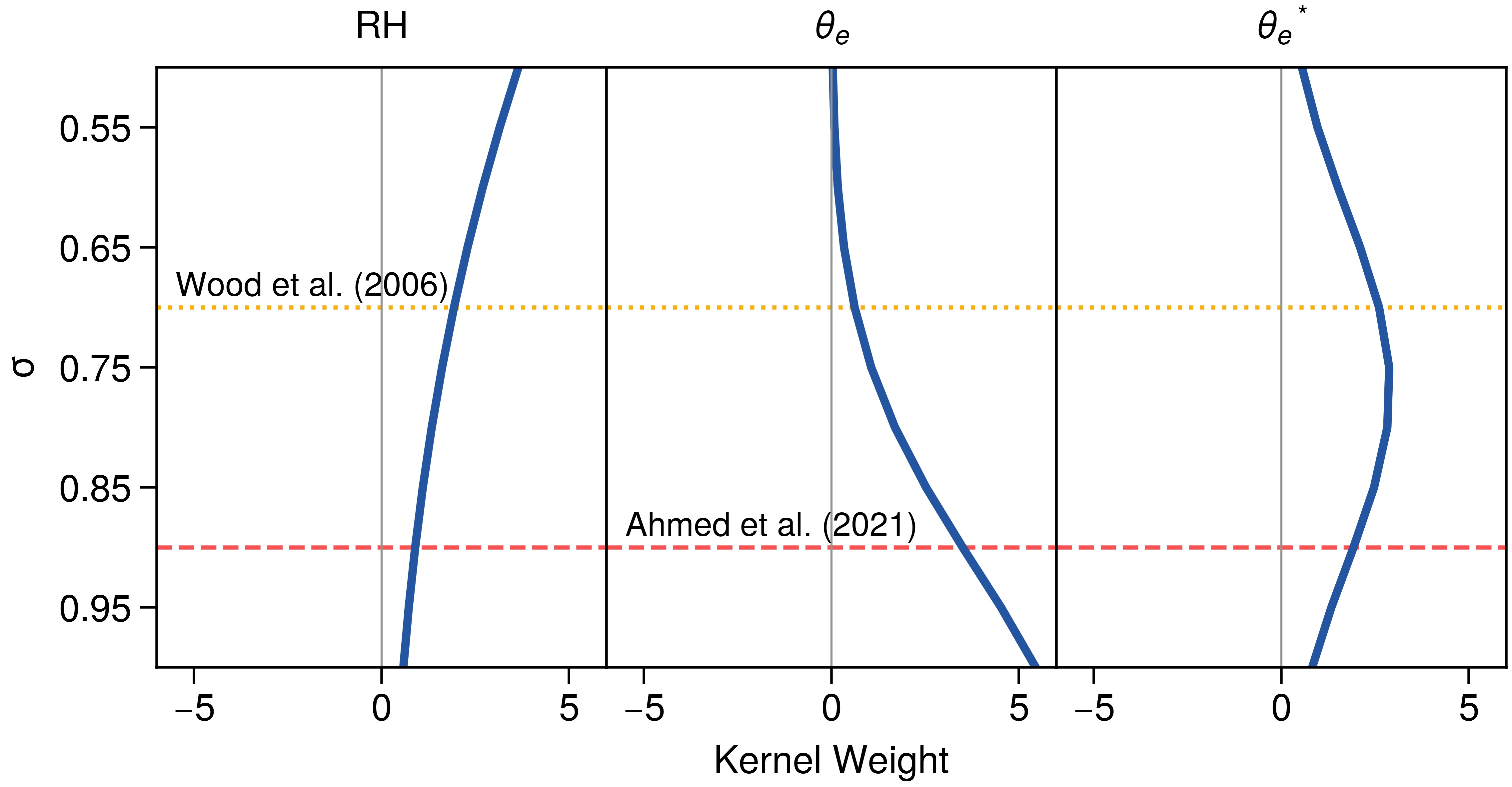}
    \caption{Gaussian kernels learned by NN-GAUSS for $\mathrm{RH}$, $\theta_e$, and $\theta_e^*$. Blue lines show mean kernel weights across seeds (\textit{n} = 3), with the extremely small inter-seed variability omitted. Absolute magnitude indicates the relative importance of each sigma level. Horizontal lines show the approximate sigma levels of commonly prescribed vertical boundaries, computed for a surface pressure of 1{,}000~hPa. The red dashed line marks the top of the 100-hPa-deep boundary layer used to compute $B_L$ \cite{Ahmed2021}, and the yellow dotted line marks the 700~hPa reference level used to compute estimated inversion strength \cite{Wood2006}.}
    \label{fig:3}
\end{figure}

Figure~\ref{fig:3} shows the Gaussian kernels learned by NN-GAUSS for $\mathrm{RH}$, $\theta_e$, and $\theta_e^*$, which define the kernel-integrated features supplied to PySR. For comparison, horizontal lines mark the approximate top of the boundary layer used to compute $B_L$ \cite{Ahmed2021} and the 700~hPa reference level used to compute estimated inversion strength \cite{Wood2006}. The kernels differ markedly across the three variables. The $\mathrm{RH}$ kernel is broad and increases toward the top of the column ($\sigma = 0.5$), consistent with dry-intrusion and entrainment controls on convective persistence in this region \cite{Menon2022, Parker2016, Volonte2020}. Because its weight is largest at the column top, moisture above 500~hPa, which we exclude for consistency with $B_L$, may also contribute to this signal. The $\theta_e$ kernel is concentrated at the surface ($\sigma = 1.0$), consistent with the role of near-surface parcel energy in convective initiation, but its weight extends well above the boundary layer over which $B_L$ averages $\theta_e$. The $\theta_e^*$ kernel peaks at $\sigma = 0.75$ in the lower free troposphere, where environmental saturation conditions determine whether rising parcels sustain vertical acceleration or detrain.

This separation between parcel energy near the surface and environmental saturation aloft is consistent with deep-inflow conceptual frameworks \cite{Ahmed2020, Schiro2018}, although the learned vertical extents differ from those prescribed in $B_L$. Integrating each thermodynamic profile with its learned kernel yields a single feature for symbolic regression. The kernels were optimized for predictive accuracy, without constraints on how precipitation should respond to those features. We therefore examine whether the resulting relationships follow the directions expected from convective physics.

\subsection{Compliance with Physical Constraints}\label{subsec:constraints}

Following the approach of \citeA{Grundner2024}, we express the expected responses of precipitation to the kernel-integrated features as three physical constraints (PCs), each a partial derivative taken with the other two features held fixed:
\begin{itemize}
    \item PC\textsubscript{1}: $\partial\mathrm{P}/\partial\widehat{\mathrm{RH}} \ge 0$
    \item PC\textsubscript{2}: $\partial\mathrm{P}/\partial\widehat{\theta_e} \ge 0$
    \item PC\textsubscript{3}: $\partial\mathrm{P}/\partial\widehat{\theta_e^*} \le 0$
\end{itemize}
Higher lower-tropospheric relative humidity reduces the dilution of rising parcels by entrained dry air, favoring precipitation (PC\textsubscript{1}). Higher parcel $\theta_e$ increases buoyancy, also favoring precipitation (PC\textsubscript{2}). At fixed parcel $\theta_e$, higher environmental $\theta_e^*$ reduces the parcel's thermodynamic advantage over its surroundings and thus tends to suppress convection (PC\textsubscript{3}).

\begin{table}[h!]
\centering
\begin{tabular}{cccc}
    \toprule
    Source & PC\textsubscript{1}: $\partial\mathrm{P}/\partial\widehat{\mathrm{RH}} \ge 0$ & PC\textsubscript{2}: $\partial\mathrm{P}/\partial\widehat{\theta_e} \ge 0$ & PC\textsubscript{3}: $\partial\mathrm{P}/\partial\widehat{\theta_e^*} \le 0$ \\
    \midrule
    ERA5 & 98.6\% & 100.0\% & 100.0\% \\
    SR-ATM & 100.0\% & 100.0\% & 100.0\% \\
    SR-SFC & 100.0\% & 100.0\% & 100.0\% \\
    SR-ALL & 100.0\% & 88.9\% & 100.0\% \\
    SR-ALL-PC & 100.0\% & 100.0\% & 100.0\% \\
    \bottomrule
    \end{tabular}
    \vspace{1em}
    \caption{Satisfaction rates of physical constraints PC\textsubscript{1}--PC\textsubscript{3} on the test set. ERA5 rates are the fraction of hypercubes whose fitted precipitation slope has the expected sign, averaged across the number of intervals per conditioning feature ($n \in \{3,4,5,6,7\}$). Symbolic regression rates are the fraction of test set samples at which the analytical partial derivatives have the expected sign. PC\textsubscript{1} is evaluated over land ($\mathrm{LF} \geq 0.5$), and PC\textsubscript{2} and PC\textsubscript{3} over all grid points. SR-BL is not included because it does not use the kernel-integrated features on which these constraints are defined.}
    \label{tab:1}
\end{table}

Table~\ref{tab:1} reports how often each constraint is satisfied in ERA5 and in the symbolic regression equations. For ERA5, we use the hypercube binning approach of \citeA{Grundner2024}. For each constraint, we divide the test set into hypercubes within which the other two kernel-integrated features are approximately constant, regress ERA5 precipitation on the tested feature within each hypercube, and compute the fraction of hypercubes whose slope has the expected sign. We repeat this for several hypercube sizes and average the results (Text~S4 and Table~S3 of Supporting Information). For the equations, we differentiate each equation analytically and report the fraction of test set samples at which each derivative has the expected sign. We evaluate the derivatives before taking the positive part in Equation~\ref{eq:prediction}, so that samples with zero predicted precipitation are not counted as automatically satisfying the constraints (Text~S4 of Supporting Information). SR-BL is not included in Table~\ref{tab:1} because it does not use the kernel-integrated features. For SR-BL, the analogous expectation is that precipitation does not decrease as $B_L$ increases ($\partial\mathrm{P}/\partial B_L \geq 0$). SR-BL satisfies this expectation for all values of $B_L$ because $\lambda$ is positive and $(B_L-B_c)^3$ increases with $B_L$ (Equation~\ref{eq:srbl}), whereas ERA5 satisfies it in 88.3\% of $B_L$ intervals (Table~S3 of Supporting Information).

In ERA5, PC\textsubscript{2} and PC\textsubscript{3} hold in 100.0\% of hypercubes over all grid points. PC\textsubscript{1} holds in 98.6\% of land hypercubes ($\mathrm{LF} \geq 0.5$), but in only 61.2\% of ocean hypercubes (Table~S3 of Supporting Information). The ocean violations occur only in hypercubes where precipitation is frequently zero and the fitted slope is near zero, so they reflect a weak response rather than a physical decrease in precipitation with moisture (Figure~S2 of Supporting Information). We therefore evaluate PC\textsubscript{1} over land and PC\textsubscript{2} and PC\textsubscript{3} over all grid points, for both ERA5 and the equations. SR-ATM, SR-SFC, and SR-ALL-PC satisfy all three constraints at every test set sample, and their derivatives show that this holds for all input values. SR-ALL satisfies PC\textsubscript{1} and PC\textsubscript{3} everywhere, but PC\textsubscript{2} at only 88.9\% of samples. We trace this violation to the form of SR-ALL, and describe the modification that yields SR-ALL-PC, in Section~\ref{subsec:srall}. The 100.0\% rates for the equations do not imply that they represent these relationships more faithfully than ERA5, since the ERA5 rates are estimated from slopes fit to finite samples, whereas the equation rates follow exactly from their analytical derivatives.

The broad agreement between ERA5 and the discovered equations is notable because none of these responses were prescribed. The PCs were not imposed during kernel learning, the symbolic regression search, or constant optimization, and the kernels were trained only to minimize prediction error. That physically consistent responses nonetheless emerge suggests that the learned kernels capture physically meaningful aspects of lower-tropospheric thermodynamic structure, although these sign constraints are necessary rather than sufficient conditions for a physical interpretation of the discovered equations.

\subsection{Competition Between Moisture and Instability}\label{subsec:sratm}

Returning to the discovered equations, SR-ATM replaces $B_L$ with the kernel-integrated features. This yields a more explicit decomposition of the thermodynamic dependence of precipitation, taking the form:
\begin{equation}
    \begin{gathered}
    \mathrm{P} = 
    \left(\exp\left(\lambda\max(M,I)^3\right)-1\right)^+,
    \text{ where} \\
    M = \kappa(\widehat{\mathrm{RH}}-\mu_\mathrm{RH})
    \text{ and }
    I = \widehat{\theta_e}-\gamma\widehat{\theta_e^*}-\Theta_c,
    \end{gathered}
    \label{eq:sratm}
\end{equation}
in which $\mu_\mathrm{RH}$ is the training mean of $\mathrm{RH}$ over all sigma levels, $\kappa$ scales $M$ to the units of $I$, $\gamma$ sets the weight of $\widehat{\theta_e^*}$ relative to $\widehat{\theta_e}$, $\Theta_c$ sets the threshold that $\widehat{\theta_e}-\gamma\widehat{\theta_e^*}$ must exceed for $I$ to become positive, and $\lambda$ controls how rapidly precipitation intensifies once the larger of the two terms becomes positive (Table~S2 of Supporting Information).

\begin{figure}[h!]
    \centering
    \includegraphics[width=\textwidth]{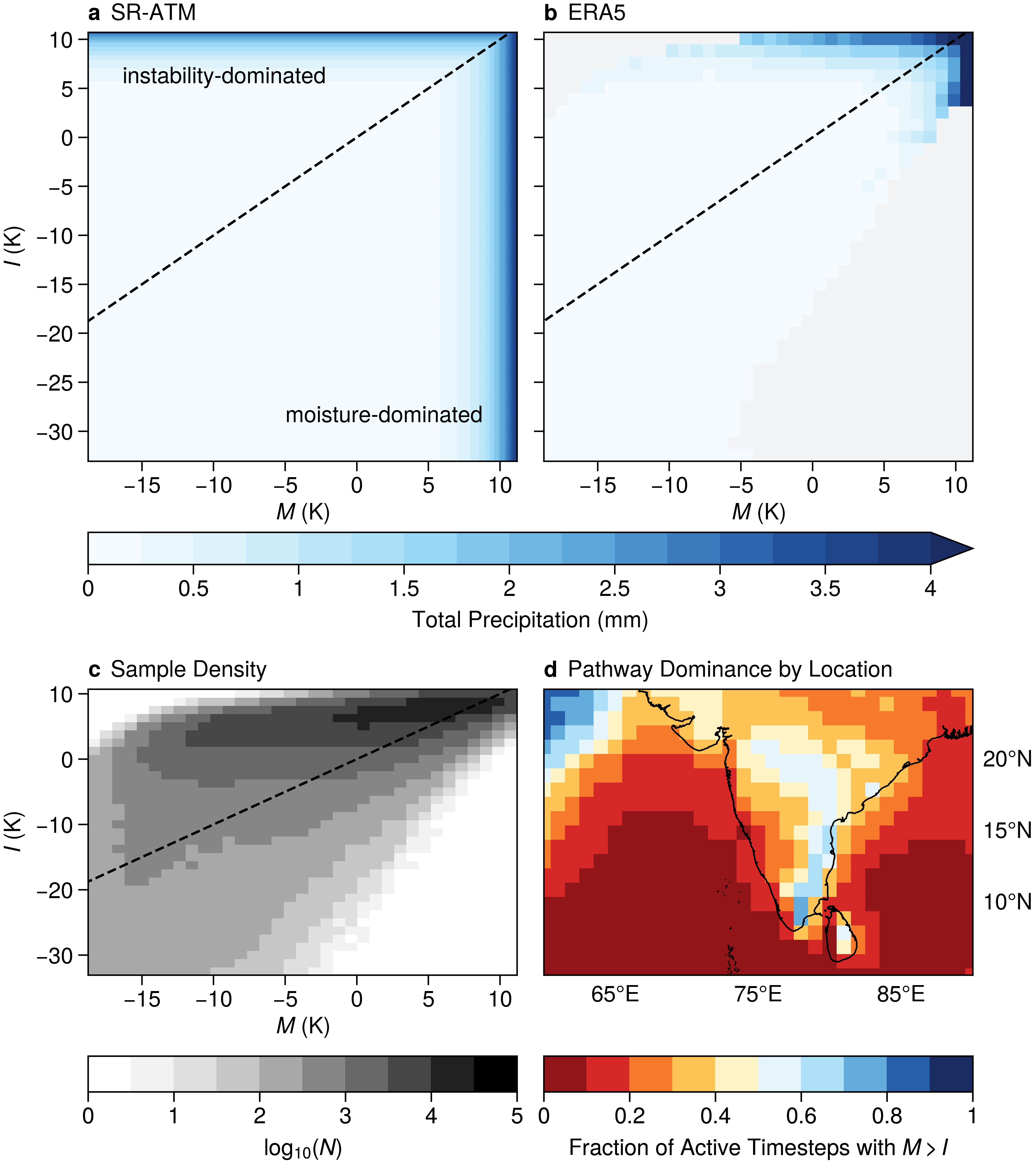} 
    \caption{Competitive thermodynamic pathways learned by SR-ATM in $M$--$I$ space, where $M$ and $I$ are the moisture and instability pathways, respectively (Equation~\ref{eq:sratm}). (a) SR-ATM predicted and (b) ERA5 bin-mean 3-hourly total precipitation in the test set, with (b) restricted to bins containing at least 50 samples and other bins shaded gray. (c) Sample density, expressed as $\log_{10}\left(N\right)$, where $N$ is the number of samples per bin. The black dashed line in (a)--(c) marks $M = I$, the boundary between moisture-dominated ($M > I$) and instability-dominated ($M < I$) conditions. (d) Fraction of active timesteps ($\max(M,I)>0$) with $M>I$ at each location, where cool colors indicate more frequent moisture dominance and warm colors indicate more frequent instability dominance.}
    \label{fig:4}
\end{figure}

Equation~\ref{eq:sratm} separates the thermodynamic dependence of precipitation into two competing pathways, one depending on lower-tropospheric moisture ($M$) and the other on a buoyancy-like instability contrast ($I$). Larger values of $\widehat{\theta_e}$ favor convection by increasing parcel energy, whereas larger values of $\widehat{\theta_e^*}$ increase the environmental saturation deficit opposing convective deepening. With $\gamma$ close to one, the instability pathway weights $\widehat{\theta_e}$ and $\widehat{\theta_e^*}$ nearly equally, resembling the thermodynamic competition underlying $B_L$ and deep-inflow frameworks \cite{Ahmed2020,Ahmed2021,Schiro2018}, although because $\gamma$ is slightly greater than one, the contrast between $\widehat{\theta_e}$ and $\widehat{\theta_e^*}$ needed for $I$ to become positive increases with $\widehat{\theta_e^*}$. Unlike these formulations, however, SR-ATM does not combine moisture and instability linearly. Instead, the $\max(\cdot)$ operator allows precipitation once either pathway exceeds its threshold, after which precipitation intensifies with the cube of the dominant pathway. Refitting Equation~\ref{eq:sratm} with additive pathways, $(aM+I)^3$ or $(aM)^3+I^3$ with an additional fitted weight $a$, lowers test R\textsuperscript{2} from 0.32 to 0.27 and 0.31, respectively. The competitive structure therefore fits the data at least as well as these additive alternatives with one fewer parameter, and it was recovered independently across all three symbolic regression seeds (Text~S3 of Supporting Information). Because $\lambda$, $\kappa$, and $\gamma$ are positive, this structure also guarantees that SR-ATM satisfies all three physical constraints (Table~\ref{tab:1}).

Figure~\ref{fig:4} examines this structure in $M$--$I$ space, where the boundary $M = I$ separates moisture-dominated ($M > I$) from instability-dominated ($M < I$) states. Figures~\ref{fig:4}a--b compare SR-ATM and ERA5 precipitation across this space. Both exhibit a similar organization, with the largest precipitation at large $M$, near and within the moisture-dominated regime, and comparatively weak precipitation in instability-dominated states. Figure~\ref{fig:4}c shows that instability-dominated states occur more frequently, despite the stronger precipitation associated with moisture-dominated states. The geographic distribution of these regimes during active conditions ($\max(M,I) > 0$) is mapped in Figure~\ref{fig:4}d. Moisture-dominated conditions (cool colors) occur most frequently over the northern Arabian Sea and the southern Coromandel Coast, whereas instability-dominated conditions (warm colors) are more frequent across the Deccan Plateau, northwestern India, and much of the Arabian Sea and Bay of Bengal. In both moisture-dominated regions, $M$ likely exceeds $I$ because $I$ is suppressed by warm, dry air descending into the lower free troposphere, which may be associated with the Pakistan heat low over the northern Arabian Sea \cite{Bollasina2011} and with westerlies descending in the lee of the Western Ghats along the southern Coromandel Coast. Along the Godavari River valley, by contrast, Bay of Bengal moisture carried inland along the monsoon trough by low-pressure systems \cite{Krishnamurthy2010} may make $M$ large, but $I$ is also large, so neither pathway consistently dominates.

These results suggest that the instability pathway primarily marks whether conditions are thermodynamically favorable for convection, consistent with SR-BL, whereas lower-tropospheric moisture distinguishes whether convection can deepen into strongly precipitating states. However, near the boundary $M = I$, ERA5 precipitation increases as both $M$ and $I$ become large, whereas SR-ATM responds only to the larger of the two, limiting its ability to capture the heaviest precipitation. SR-ATM also depends only on the atmospheric state, whereas the partitioning of surface energy between boundary layer moistening and heating could shift the balance between the two pathways. We therefore next test whether surface conditions account for the errors left by SR-ATM.

\subsection{Surface Corrections Independent of the Atmospheric State}\label{subsec:srsfc}

The learned equation SR-SFC extends SR-ATM by allowing surface conditions to modify the precipitation predicted from the atmospheric thermodynamic state:
\begin{equation}
    \begin{gathered}
    \mathrm{P} = \left(\exp\left(\lambda\max(M,I)^3+S\right)-1\right)^+,
    \text{ where} \\
    S = \lambda_\mathrm{SHF}(\mathrm{LF}_c-\mathrm{LF})(\mathrm{SHF}-\mu_\mathrm{SHF})+\lambda_\mathrm{LHF}(\mathrm{LHF}-\mu_\mathrm{LHF}),
    \end{gathered}
    \label{eq:srsfc}
\end{equation}
in which $M$, $I$, and $\lambda$ are unchanged from SR-ATM (Equation~\ref{eq:sratm}), $\mu_\mathrm{SHF}$ and $\mu_\mathrm{LHF}$ are the training means of $\mathrm{SHF}$ and $\mathrm{LHF}$, $\lambda_\mathrm{SHF}$ and $\lambda_\mathrm{LHF}$ scale the response of precipitation to departures of each flux from its mean, and $\mathrm{LF}_c$ is the land fraction at which the $\mathrm{SHF}$ response reverses sign (Table~S2 of Supporting Information).

The surface term $S$ acts as a correction to SR-ATM, increasing the SR-ATM precipitation prediction when $S > 0$ and reducing it when $S < 0$. Because $S$ enters the exponent alongside the atmospheric term ($\lambda\max(M,I)^3$), it primarily modifies precipitation where the thermodynamic state already supports convection. Its dependence on $\mathrm{LF}$ allows the surface fluxes to affect this correction differently over ocean and land. Because $S$ contains no kernel-integrated features, SR-SFC retains the thermodynamic responses of SR-ATM and satisfies all three physical constraints (Table~\ref{tab:1}).

\begin{figure}[h!]
    \centering
    \includegraphics[width=\textwidth]{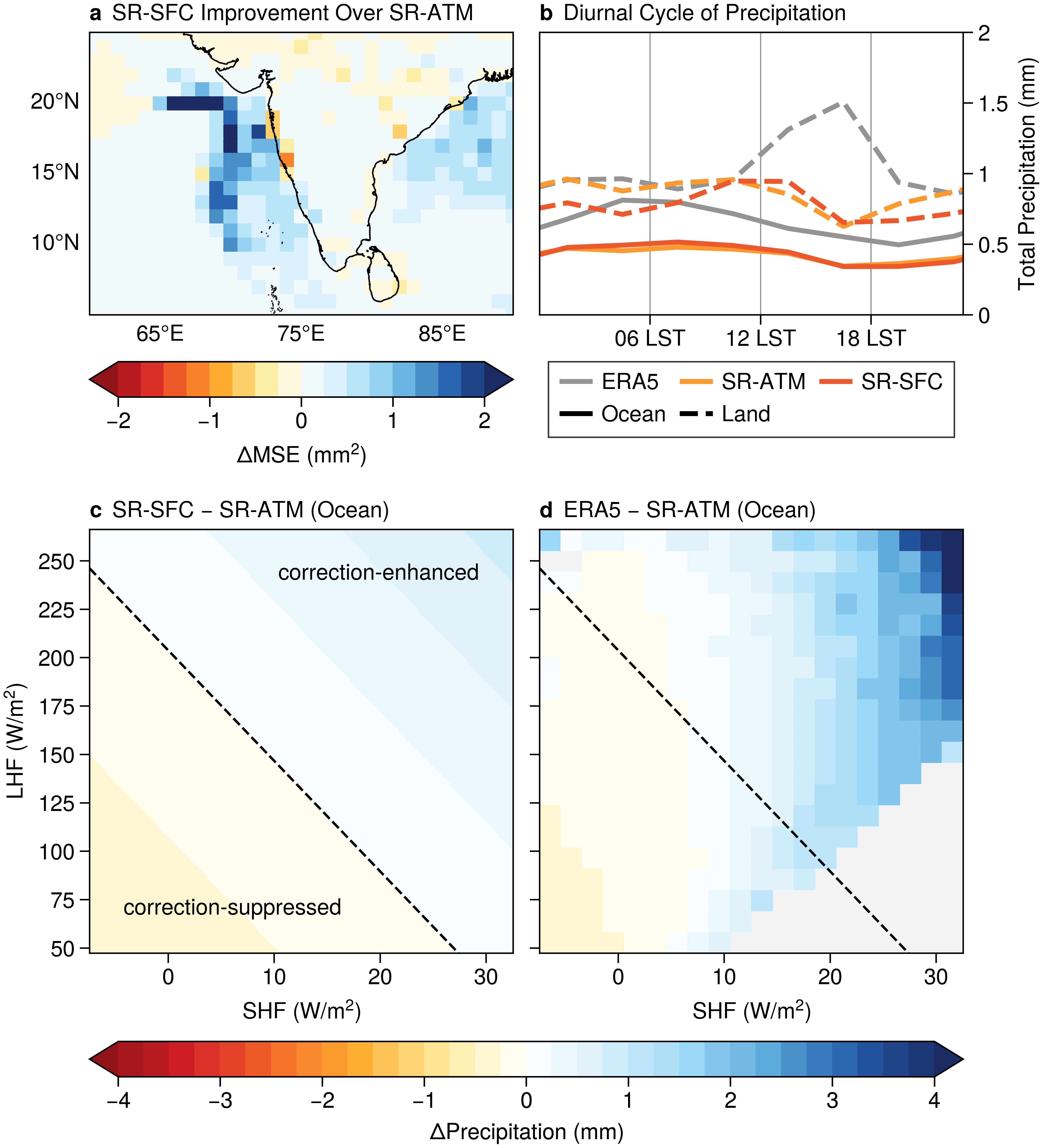} 
    \caption{Precipitation correction learned by SR-SFC (Equation~\ref{eq:srsfc}). (a) Change in test set MSE relative to SR-ATM at each location, where cool colors indicate that SR-SFC reduces MSE and warm colors indicate that it increases MSE. (b) Mean 3-hourly total precipitation from ERA5, SR-ATM, and SR-SFC over ocean ($\mathrm{LF} < 0.5$, solid) and land ($\mathrm{LF} \geq 0.5$, dashed), as a function of local solar time at the middle of each 3-hour window. (c) SR-SFC predicted and (d) ERA5 bin-mean differences in 3-hourly total precipitation from SR-ATM over ocean in $\mathrm{SHF}$--$\mathrm{LHF}$ space, for test set samples in which SR-ATM predicts at least 0.1~mm, with (c) averaged over the land fractions and SR-ATM precipitation of these samples and (d) restricted to bins containing at least 50 samples, with other bins shaded gray. In (c)--(d), cool colors indicate more precipitation than SR-ATM and warm colors indicate less, and the black dashed line marks where the correction in (c) is zero.}
    \label{fig:5}
\end{figure}

Figure~\ref{fig:5} examines where and how the surface correction modifies the predictions of SR-ATM. Figure~\ref{fig:5}a shows that the largest improvements occur over the eastern Arabian Sea, with smaller improvements across parts of the Bay of Bengal and little change over most land regions apart from a few degraded coastal grid points. Consistent with this spatial pattern, ocean R\textsuperscript{2} increases from approximately 0.34 to 0.43, whereas land R\textsuperscript{2} changes little. Figure~\ref{fig:5}b shows that the surface correction also affects the diurnal cycle differently over ocean and land. Over ocean (solid lines), SR-SFC modestly increases morning precipitation and strengthens the diurnal variation, raising the difference between its highest and lowest 3-hourly values from 42\% to 55\% of that in ERA5. Precipitation nonetheless remains underestimated throughout the day, partly because of the log-space training objective (Section~\ref{subsubsec:framework}). Over land (dashed lines), the more notable effect is on timing. SR-SFC shifts the precipitation maximum from around midnight to late morning, closer to the late-afternoon maximum in ERA5, although it still substantially underestimates this daytime maximum and reduces nighttime precipitation below ERA5 (Section~\ref{sec:conclusion}).

Because the largest skill improvements occur over ocean, Figures~\ref{fig:5}c--d examine the surface correction over ocean in $\mathrm{SHF}$--$\mathrm{LHF}$ space. SR-SFC broadly captures the structure in ERA5, increasing precipitation under strong sensible and latent heat fluxes and suppressing it under weak ones. However, ERA5 indicates that this relationship depends primarily on sensible heating, with precipitation increasingly exceeding SR-ATM as $\mathrm{SHF}$ increases and $\mathrm{LHF}$ mattering mainly when $\mathrm{SHF}$ is large. SR-SFC instead gives both fluxes substantial influence at all flux values and strongly underestimates the precipitation enhancement associated with strong sensible heating.

These results suggest that surface conditions contain information about precipitation beyond the atmospheric thermodynamic state alone, although their relationship with precipitation differs between ocean and land. Over ocean, SR-SFC captures a relationship between stronger sensible heating and precipitation underestimated by SR-ATM, producing substantial regional skill gains. Over land, the correction adds little overall skill but shifts precipitation toward daytime, when surface heating is stronger. However, SR-SFC gives $\mathrm{LHF}$ too much influence over ocean and responds too weakly to large $\mathrm{SHF}$. Because the surface fluxes and precipitation are concurrent, their relationship also does not establish whether surface heating and moistening promote precipitation, respond to it, or both. Over ocean, sensible heat flux also scales with near-surface wind speed and the air--sea temperature difference, both of which increase under strong low-level flow and convective downdrafts, so the correction may partly reflect these processes rather than surface heating alone. More fundamentally, $S$ depends only on the surface state and land fraction, so the correction itself cannot vary with the atmospheric thermodynamic state. SR-ALL relaxes this restriction by allowing its correction to depend on both the surface variables and the kernel-integrated features, which we examine next.

\subsection{Surface Corrections Coupled to the Atmospheric State}\label{subsec:srall}

SR-ALL extends SR-ATM with a correction that can draw on both the surface variables and the kernel-integrated features, taking the form:
\begin{equation}
    \begin{gathered}
    \mathrm{P} = \left(\exp\left(\lambda\max(M,I)^3+C\right)-1\right)^+, 
    \text{ where} \\
    C = (\mathrm{LF}_c-\mathrm{LF})^3\left(\lambda_{\theta_e}(\widehat{\theta_e}-\mu_{\theta_e})+\lambda_\mathrm{SHF}(\mathrm{SHF}-\mu_\mathrm{SHF})\right),
    \end{gathered}
    \label{eq:srall}
\end{equation}
in which $M$, $I$, and $\lambda$ are unchanged from SR-ATM (Equation~\ref{eq:sratm}), $\mu_{\theta_e}$ is the training mean of $\theta_e$ over all sigma levels, $\lambda_{\theta_e}$ and $\lambda_\mathrm{SHF}$ scale the response of precipitation to $\widehat{\theta_e}$ and $\mathrm{SHF}$ relative to $\mu_{\theta_e}$ and $\mu_\mathrm{SHF}$, and $\mathrm{LF}_c$ is the land fraction at which both responses reverse sign (Table~S2 of Supporting Information).

Of the six predictors available to its correction, SR-ALL uses only three, scaling $\widehat{\theta_e}$ and $\mathrm{SHF}$ by a shared land fraction term, $(\mathrm{LF}_c-\mathrm{LF})^3$. Where $\mathrm{LF} < \mathrm{LF}_c$, higher near-surface $\theta_e$ and stronger sensible heating are both associated with more precipitation than SR-ATM predicts. Because the land fraction term is cubed, the correction is about 13 times weaker at $\mathrm{LF} = 1$ than at $\mathrm{LF} = 0$, so it acts mainly over ocean and coastal grid points. Unlike $S$ in SR-SFC, $C$ depends on the thermodynamic state through $\widehat{\theta_e}$, allowing the surface correction to vary with near-surface parcel energy.

However, this dependence on $\widehat{\theta_e}$ also causes SR-ALL to violate PC\textsubscript{2} (Table~\ref{tab:1}). At 11.1\% of test set samples, SR-ALL predicts less precipitation when near-surface $\theta_e$ increases, even though higher parcel energy should favor convection. The cause is the shared land fraction term. Where $\mathrm{LF} > \mathrm{LF}_c \approx 0.70$, $(\mathrm{LF}_c-\mathrm{LF})^3$ becomes negative and reverses the sign of both terms it multiplies. For $\mathrm{SHF}$, this reversal over land resembles SR-SFC, but for $\widehat{\theta_e}$ it means that higher parcel energy is associated with less precipitation. PC\textsubscript{2} is violated wherever this negative weight outweighs the positive response of SR-ATM to $\widehat{\theta_e}$, which is zero in moisture-dominated states.

To remove this violation, we define a new model, SR-ALL-PC, by modifying SR-ALL so that it satisfies PC\textsubscript{2} for all inputs, similar to the approach of \citeA{Grundner2024}. SR-ALL-PC takes the form:
\begin{equation}
    \begin{gathered}
    \mathrm{P} = \left(\exp\left(\lambda\max(M,I)^3+C\right)-1\right)^+, 
    \text{ where} \\
    C = \lambda_{\theta_e}\left[(\mathrm{LF}_c-\mathrm{LF})^3+(1-\mathrm{LF}_c)^3\right](\widehat{\theta_e}-\mu_{\theta_e}) + \lambda_\mathrm{SHF}(\mathrm{LF}_c-\mathrm{LF})^3(\mathrm{SHF}-\mu_\mathrm{SHF}),
    \end{gathered}
    \label{eq:srallpc}
\end{equation}
in which $M$, $I$, and $\lambda$ are unchanged from SR-ATM (Equation~\ref{eq:sratm}), and the constants in $C$ are reoptimized (Table~S2 of Supporting Information). The only structural change from SR-ALL is the added $(1-\mathrm{LF}_c)^3$ in the weight on $\widehat{\theta_e}$. Without it, this weight decreases as $\mathrm{LF}$ increases and becomes negative where $\mathrm{LF} > \mathrm{LF}_c$, reaching $-\lambda_{\theta_e}(1-\mathrm{LF}_c)^3$ at $\mathrm{LF} = 1$. The added term shifts the weight upward by exactly this amount, so its lowest value, at $\mathrm{LF} = 1$, becomes zero. The weight on $\widehat{\theta_e}$ is therefore never negative, and SR-ALL-PC satisfies PC\textsubscript{2} for all input values (Table~\ref{tab:1}). This modification does not reduce predictive skill, with SR-ALL and SR-ALL-PC both achieving a test R\textsuperscript{2} of approximately 0.42 (Figure~\ref{fig:1}).

\begin{figure}[h!]
    \centering
    \includegraphics[width=\textwidth]{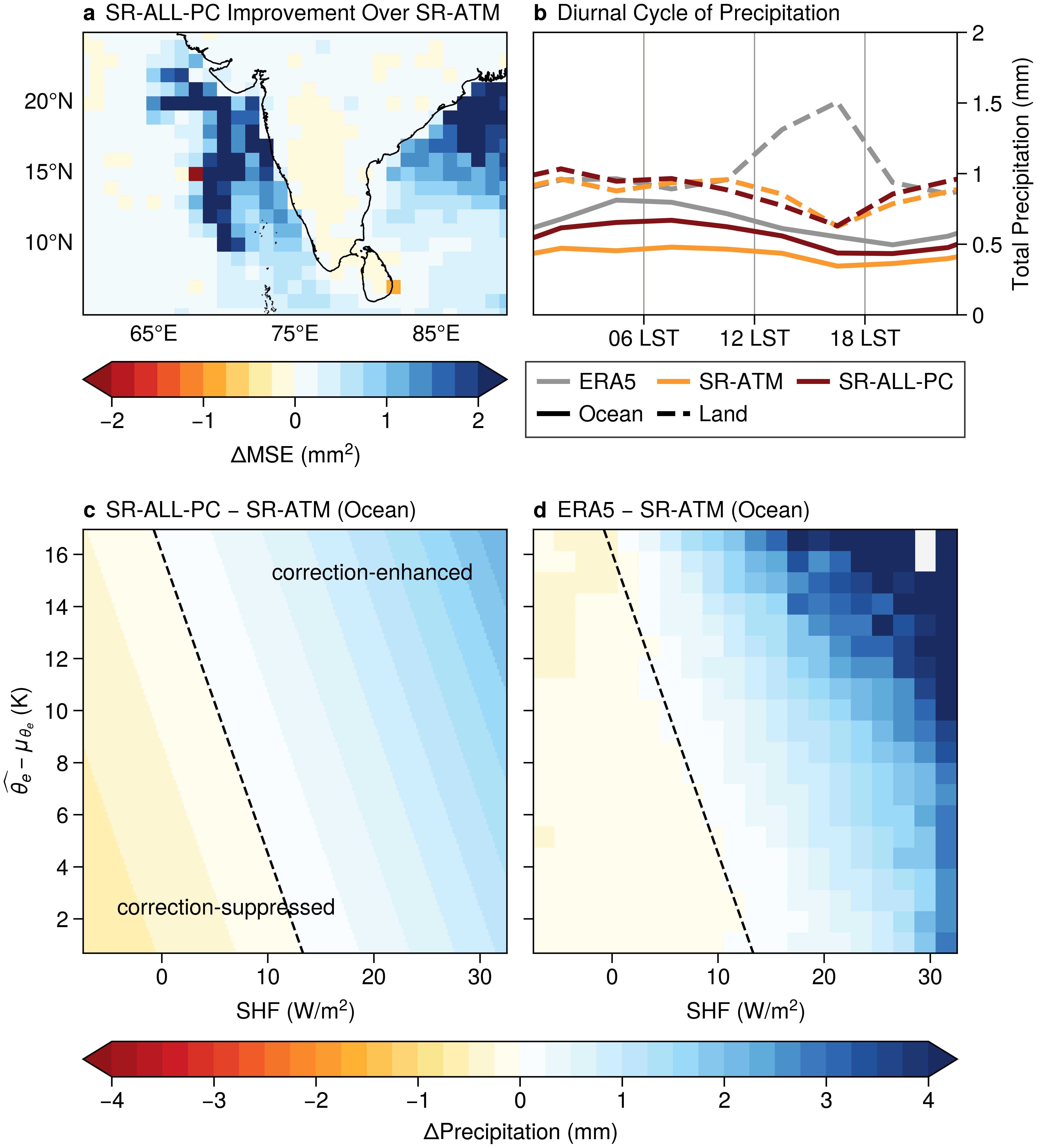} 
    \caption{As in Figure~\ref{fig:5}, but for the precipitation correction learned by SR-ALL-PC (Equation~\ref{eq:srallpc}), with (c)--(d) shown in $\mathrm{SHF}$--$(\widehat{\theta_e}-\mu_{\theta_e})$ space and the black dashed line marking where the correction in (c) is zero.}
    \label{fig:6}
\end{figure}

Figure~\ref{fig:6} examines how coupling sensible heating to the atmospheric thermodynamic state further modifies the predictions of SR-ATM. Figure~\ref{fig:6}a shows the same spatial pattern of improvement seen with SR-SFC, but with substantially larger reductions in MSE over the eastern Arabian Sea and the northern Bay of Bengal. Over land, skill changes little. Consistent with this contrast, ocean R\textsuperscript{2} increases from approximately 0.34 in SR-ATM to 0.52 in SR-ALL-PC, whereas land R\textsuperscript{2} remains nearly unchanged. Figure~\ref{fig:6}b shows that over ocean (solid lines), SR-ALL-PC increases precipitation throughout the diurnal cycle, particularly in the early morning when ERA5 precipitation peaks, reducing the persistent underprediction in SR-ATM. This substantially improves the amplitude of the oceanic diurnal cycle, which reaches 74\% of the ERA5 range compared with 42\% for SR-ATM, although precipitation remains underestimated, as for SR-SFC. Over land (dashed lines), SR-ALL-PC instead slightly increases nighttime precipitation and reduces daytime precipitation relative to SR-ATM. Where $\mathrm{LF} > \mathrm{LF}_c$, the weight on $\mathrm{SHF}$ becomes negative, so strong daytime sensible heating is associated with less precipitation, possibly because clouds and precipitation reduce $\mathrm{SHF}$ over land. This moves the land diurnal cycle further from ERA5, unlike the improved timing produced by SR-SFC.

Because the improvement again occurs primarily over ocean, Figures~\ref{fig:6}c--d examine the correction over ocean in $\mathrm{SHF}$--$(\widehat{\theta_e}-\mu_{\theta_e})$ space. SR-ALL-PC increases precipitation as sensible heating and near-surface parcel energy increase, while suppressing precipitation when both are weak. ERA5 shows a similar dependence, with precipitation increasingly exceeding SR-ATM as both $\mathrm{SHF}$ and $\widehat{\theta_e}$ increase. Unlike in Figure~\ref{fig:5}, ERA5 changes sign close to the dashed line, indicating that SR-ALL-PC correctly identifies the conditions under which SR-ATM underestimates or overestimates oceanic precipitation. This structure suggests that the relationship between sensible heating and precipitation depends on the thermodynamic state of the lower atmosphere, with its strongest association with precipitation occurring when near-surface parcel energy is also high. SR-ALL-PC captures this coupled response more directly than SR-SFC, although it still underestimates the magnitude of the precipitation enhancement when both $\mathrm{SHF}$ and $\widehat{\theta_e}$ are large.

These results suggest that the oceanic relationship between surface fluxes and precipitation identified by SR-SFC is better represented by coupling sensible heating to the atmospheric thermodynamic state than by treating surface fluxes independently of it. This coupling substantially improves predictions of oceanic precipitation and its diurnal variability while satisfying all three physical constraints. The nearly identical performance of SR-ALL and SR-ALL-PC further shows that the unphysical response to $\widehat{\theta_e}$ can be removed without sacrificing predictive skill. However, SR-ALL-PC still responds too weakly when sensible heating and near-surface parcel energy are both large, and its reversed response to $\mathrm{SHF}$ over land shifts precipitation away from the daytime maximum. Nonetheless, the final equation identifies a coupled oceanic relationship in which sensible heating becomes increasingly associated with precipitation in thermodynamically favorable environments.

\section{Conclusion}\label{sec:conclusion}

While the thermodynamic ingredients relevant to South Asian monsoon precipitation are well recognized, how they combine to shape precipitation across the domain remains unclear. A widely used diagnostic of lower-tropospheric buoyancy summarizes the thermodynamic environment as a balance between convective instability and lower free-tropospheric dryness \cite{Ahmed2021}, but imposes strong assumptions about both the vertical structure of thermodynamic influence and the functional form relating that structure to precipitation. In this study, we use a hierarchy of data-driven models to examine how much these imposed structures limit predictive skill and what can be learned when they are relaxed. Models given this buoyancy diagnostic alone capture substantially less precipitation variability than models with access to surface fluxes and the resolved lower-tropospheric thermodynamic state, indicating that the diagnostic does not retain all of the relevant information. The most skillful of these models are neural networks, whose main disadvantage is that their many parameters obscure physical interpretation. Distilling learned representations of vertical thermodynamic structure through symbolic regression yields compact, closed-form equations in which both the vertical structure and the functional form emerge directly from the data. The best of these equations recovers about 85\% of the skill of the kernel-based neural networks and over 70\% of that of the full neural network, with roughly three orders of magnitude fewer parameters.

These equations reveal a hierarchy of thermodynamic dependencies built on buoyancy. Buoyancy serves as a baseline predictor, capturing the transition from suppressed to active convection, with precipitation increasing progressively faster as conditions become more favorable, but it captures only part of the precipitation variability. When the moisture and instability that the buoyancy diagnostic combines into a single quantity are instead learned separately, they act as competing pathways rather than additive contributions. Instability primarily marks whether conditions are thermodynamically favorable for convection, whereas lower-tropospheric moisture distinguishes whether convection reaches heavily precipitating states. Including surface conditions reveals a coupling between the surface and the lower atmosphere. Over ocean, precipitation is associated not with sensible heating alone but with its combination with near-surface equivalent potential temperature, with the most precipitation occurring when both are high. Each of these dependencies improves predictive skill, and the equations largely agree with expected physical constraints that were not imposed during learning, with the one violation removed by a simple modification. Together, these relationships help reconcile longstanding moisture- and buoyancy-centered perspectives by framing them not as competing explanations but as components of a hierarchical system.

The improvements from adding surface conditions are, however, concentrated over ocean. Over land, the kernel-integrated features improve skill relative to SR-BL, but skill changes little once surface variables are added, even as the equations become more complex, and the equations represent the diurnal cycle of precipitation poorly. This may reflect thermodynamic behavior that is difficult to express in a compact equation relating precipitation only to the concurrent state. Afternoon convection over land develops as near-surface heating and moistening build through the morning \cite{Dai2001}, and this heating also drives low-level convergence toward elevated terrain such as the Deccan Plateau \cite{Chen2020}, so the same thermodynamic state can be associated with different precipitation depending on the time of day. Even a mesoscale model with the full atmospheric state and parameterized convection peaks too early and underestimates the diurnal amplitude over central India \cite{Bhate2019}, so this limitation is not unique to the discovered equations. 

Skill could improve further by reoptimizing the kernels jointly with the equation constants \cite{Lemos2023}, as the integration kernel parameters were learned within the neural network NN-GAUSS rather than for the discovered equations. Beyond what either the equations or the neural networks capture, the remaining unexplained variability indicates that concurrent thermodynamic structure alone cannot fully account for monsoon precipitation. This is especially true across South Asia, where complex topography makes the location and intensity of convection depend strongly on how the large-scale flow interacts with the terrain, for instance through dry intrusions \cite{Deoras2026}. Extending the same framework to include a few learned features of the dynamical state, such as low-level winds or convergence, could test how much of this remaining variability can be explained while retaining compact, interpretable equations.


\acknowledgments

We thank Ariane Bietighoefer, Spencer Hill, Jerry Lin, Sara Shamekh, and Frederick Iat-Hin Tam for valuable scientific discussions that helped inform this work. S.L.F. was supported by the National Science Foundation (NSF) Science and Technology Center (STC) Learning the Earth with Artificial Intelligence and Physics (LEAP) (2019625-STC) and the U.S. Department of Energy (DOE) Advanced Scientific Computing Research (ASCR) Program (DE-SC0022255). M.S.P. acknowledges support from NVIDIA Corporation. J.W.B. was supported by the National Aeronautics and Space Administration (NASA) New Investigator Program in Earth Science (NIP) (80NSSC21K1735). T.B. was supported by the Swiss State Secretariat for Education, Research, and Innovation (SERI) under the Horizon Europe AI4PEX Project (Grant Agreement ID 101137682; SERI No. 23.00546). Computational resources were provided by Perlmutter at the National Energy Research Scientific Computing Center (NERSC), a DOE User Facility (Project m4334).

\section*{Conflict of Interest}

The authors declare no conflicts of interest relevant to this study.

\section*{Availability Statement}

The ERA5 reanalysis data used to derive our predictors and target
\cite{Hersbach2023a, Hersbach2023b} are available as an analysis-ready, cloud-optimized version via the LEAP Data Catalog (\url{https://catalog.leap.columbia.edu/feedstock/arco-era5}). The code used for data processing, model training and inference, and visualization, together with the Pareto frontiers of every symbolic regression search and seed, is archived on Zenodo (\url{https://doi.org/10.5281/zenodo.24094998}).


\bibliography{references}

\appendix
\section*{Supporting Information}

\setcounter{figure}{0}
\setcounter{table}{0}
\setcounter{equation}{0}
\renewcommand{\thefigure}{S\arabic{figure}}
\renewcommand{\thetable}{S\arabic{table}}
\renewcommand{\theequation}{S\arabic{equation}}

\noindent\textbf{Text S1. Recovering Physical Precipitation}

This text shows how the model output $z$ maps to physical precipitation in Equation~2 (Section~2.2.1). The standardized value of zero precipitation is $-\mu_y/s_y$, so the standardized prediction is $-\mu_y/s_y+z^+$. Inverting the standardization ($y=\mu_y+s_y[-\mu_y/s_y+z^+]$) and then the log-transformation ($\mathrm{P}=\exp(y)-1$) gives:
\begin{equation*}
    \mathrm{P} 
    = \exp\left(\mu_y+s_y\left[-\frac{\mu_y}{s_y}+z^+\right]\right)-1 
    = \exp(s_yz^+)-1 
    = (\exp(s_yz)-1)^+,
\end{equation*}
so $\mu_y$ cancels. The last equality holds because $s_y > 0$ and the exponential is increasing, so $\exp(s_yz^+)-1$ equals $\exp(s_yz)-1$ when $z > 0$, and equals zero when $z \leq 0$, where $\exp(s_yz)-1 \leq 0$. This gives Equation~2.

\noindent\textbf{Text S2. Gaussian Kernel Parameterization}

The Gaussian kernel formulation summarized in Section~2.2.3 represents the vertical weighting of each predictor with a learned center and width. In this text, we first define this weighting function, then use Table~S1 and Figure~S1 to examine what the Gaussian representation captures and why it is retained for symbolic regression.

\textit{Gaussian Kernel Definition.} For predictor $\varphi_i$, the continuous weighting function $k_i(\sigma)$ is defined by a learned center ($\sigma_c$) and width ($\sigma_w$) in terrain-following sigma coordinates, where $\sigma = 1$ is the surface:
\begin{equation*}
    k_i(\sigma) 
    = \exp\left(-\frac{(\sigma-\sigma_c)^2}{2{\sigma_w}^2}\right).
\end{equation*}
This continuous function is evaluated at discrete sigma levels within the column ($0.5 \leq \sigma \leq 1$) and normalized to integrate to unity during kernel integration (Equation~4). Because $\sigma_c$ is not restricted to the column, the learned Gaussian can peak above or below the column, in which case its weight increases monotonically toward the nearer edge of the column. Table~S1 therefore describes each kernel by the sigma level where its weight peaks within the column (which lies at the column edge for $\mathrm{RH}$ and $\theta_e$) and by the spread of its weight around that level.

\textit{Comparison with Nonparametric Kernels.} To assess whether this low-dimensional parameterization adequately captures the dominant vertical thermodynamic structure, Figure~S1 compares the Gaussian kernels with nonparametric kernels in which weights are learned independently at each sigma level.

The nonparametric kernels recover fine structure, including multimodal and oscillatory features, whereas the Gaussian formulation compresses each predictor into a single effective vertical scale. For $\mathrm{RH}$, the nonparametric kernel is approximately bimodal, with substantial contributions from moisture near the surface and near the top of the column, while the Gaussian representation smooths this into a single broad weighting that increases toward the top of the column. For $\theta_e$, both representations emphasize strong near-surface weighting, consistent with the role of boundary layer energy in parcel initiation. For $\theta_e^*$, both center in the lower free troposphere, where the environmental saturation state regulates convective deepening. Although the nonparametric $\theta_e^*$ kernel exhibits two distinct positive lobes, the Gaussian representation compresses these into a single lower free-tropospheric layer.

Despite these structural differences, the Gaussian kernel model (NN-GAUSS) achieves skill on the validation set similar to the nonparametric kernel model (NN-NONPARAM; R\textsuperscript{2} $\approx$ 0.51 for both) while using fewer kernel parameters. We therefore retain single Gaussian kernels for their parsimony and interpretability, which provide greater insight into the dominant vertical structure relevant to precipitation without sacrificing predictive performance.

\begin{table}[h!]
\centering
\small
\begin{tabular}{ccc}
    \toprule
    \textbf{Predictor} & \textbf{Peak Level} & \textbf{Spread} \\
    \midrule
    $\mathrm{RH}$ & $0.50$ & $0.219$ \\
    $\theta_e$ & $1.00$ & $0.139$ \\
    $\theta_e^*$ & $0.75$ & $0.125$ \\
    \bottomrule
    \end{tabular}
    \vspace{1em}
    \caption{Sigma level of maximum weight (peak level) and root second moment of the weights about that level (spread) for each Gaussian kernel, computed from kernel weights averaged across training seeds (\textit{n} = 3). Inter-seed variability is negligible and therefore omitted.}
\end{table}

\begin{figure}[h!]
    \centering
    \includegraphics[width=0.8\textwidth]{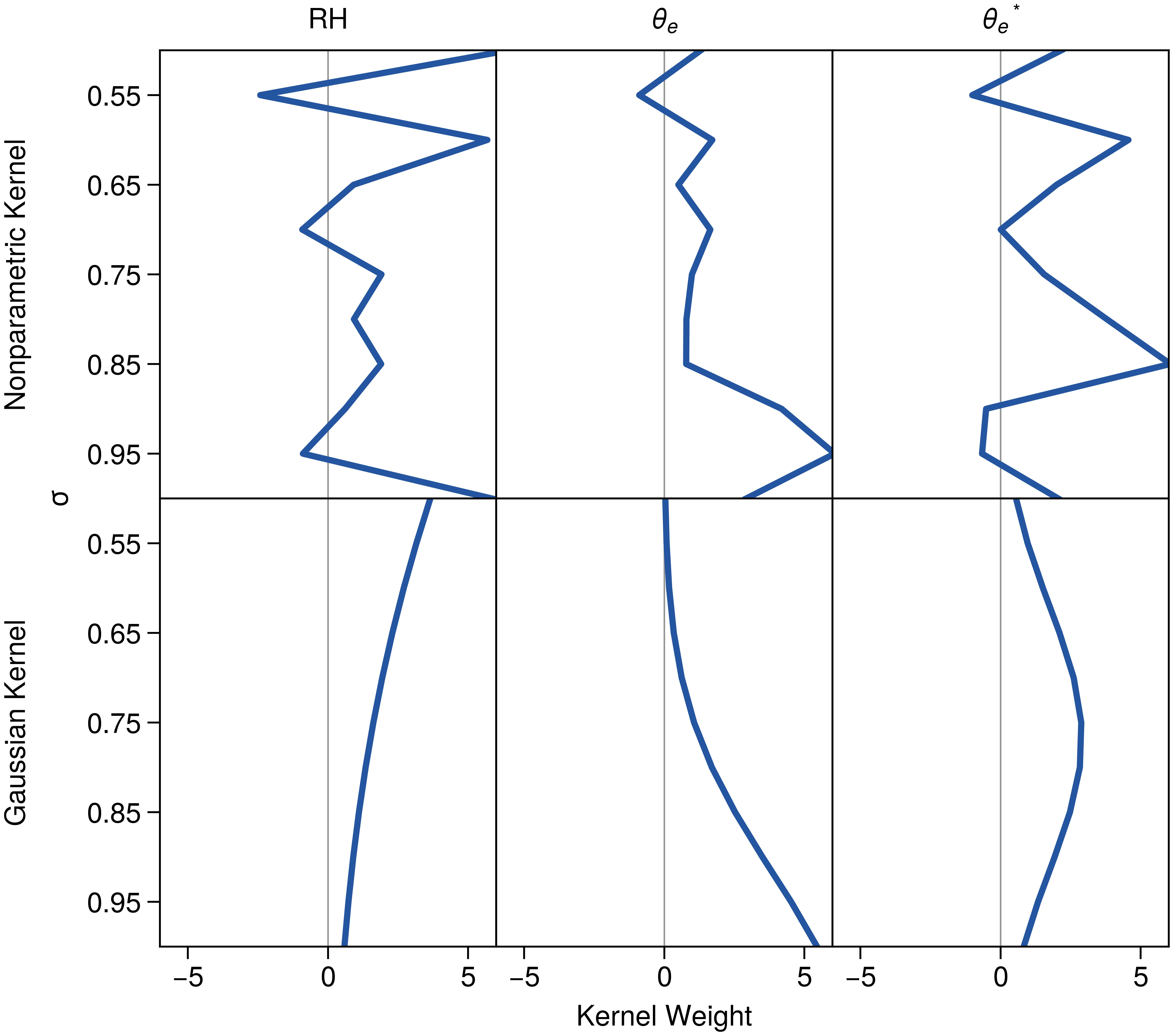}
    \caption{Learned nonparametric (top) and Gaussian (bottom) vertical kernels for $\mathrm{RH}$, $\theta_e$, and $\theta_e^*$. Blue lines show mean weights across training seeds (\textit{n} = 3), with the extremely small inter-seed variability omitted. Absolute magnitude indicates the relative importance of each sigma level.}
\end{figure}

\noindent\textbf{Text S3. Symbolic Regression Search and Optimization}

This text provides methodological details for the symbolic regression procedure summarized in Section~2.2.4. It specifies the search settings and the complexity measure used in Section~3.1, and describes how training samples were selected and how equation constants were optimized.

\textit{Search Settings and Complexity.} Searches with vertically varying predictors use kernel-integrated features computed with the NN-GAUSS kernels averaged across seeds (Figure~3). The search uses binary operators $\{+,-,\times,\div,(\cdot)^{(\cdot)},\min(\cdot),\max(\cdot)\}$ and unary operators $\{|\cdot|,(\cdot)^2,(\cdot)^3,\mathrm{neg}(\cdot),\exp(\cdot)\}$, and minimizes the same MSE objective as the neural networks, applied after taking the positive part of each candidate output (Section~2.2.1). 

Complexity is measured by weighted node count, with each predictor weighted by its information content. Constants and $\mathrm{LF}$ (which is time-invariant and bounded within $[0,1]$) contribute 1 unit each. Time-varying predictors (the kernel-integrated features and surface fluxes) contribute 2 units each, as does the output of an existing equation in searches that build on one. There are two exceptions. $B_L$ contributes 1 unit despite varying in time, because weights only set how strongly the search favors one predictor over another, and $B_L$ is the only predictor in the SR-BL search. In the SR-ALL search, the kernel-integrated features contribute 1 unit each, because with 2 units the search consistently favored corrections built from the surface variables alone. Operators $\{+,-,\times,\mathrm{neg}(\cdot)\}$ contribute 1 unit and all others 2 units, penalizing more complex operations. The maximum complexity is 20 (10 for SR-BL) and the maximum tree depth is 10. To preserve interpretability, power-law exponents and denominators are limited in complexity and selected nested compositions (e.g., an exponential inside another exponential) are disallowed. Candidate equations evolve within 20 populations of 150 members over 200 iterations, with a parsimony coefficient of 0.0025 and a constant-optimization weight of 0.25.

\textit{Subsampling Strategy.} Because the combined training and validation set (approximately 8.6 million samples) is prohibitively large for symbolic regression, each search uses a stratified subset of approximately 215,000 samples (2.5\%), drawn separately for each seed. Timesteps are grouped by domain-maximum precipitation into $\log_{10}$ bins spanning 10\textsuperscript{--4} to 10\textsuperscript{2}~mm plus a dry bin, then sampled in proportion to bin frequency. Sampling whole timesteps retains all grid points within each, preserving spatial covariance structure while avoiding geographic sampling biases. This is the smallest subset size that maintains stable estimates of key statistical properties of the full dataset, including dry fraction (approximately 19\%), central tendency, variability, and upper-tail behavior. Because each seed draws a different subset, structures that recur across seeds are also robust to the choice of subset, and the constants of each selected structure are refit on the full combined training and validation set.

\textit{Constant Optimization.} The constants of each selected equation are reoptimized on the full combined training and validation set using the same objective, with the L-BFGS-B algorithm and 50 initializations. One initialization uses the PySR constants of the selected structure, averaged across the seeds whose equation at the selected complexity has exactly that structure. These are seeds 42, 72, and 102 for SR-BL and SR-ALL, and seeds 42 and 72 for SR-ATM and SR-SFC. Seed 102 also recovered the $\max(M,I)$ structure of SR-ATM at the same complexity, but with a constant offset outside the cube instead of a threshold inside it, so its constants were not used for initialization. SR-ALL-PC, which was not discovered by PySR, is instead initialized from the optimized SR-ALL constants. The remaining 49 initializations are drawn uniformly from $[-5,5]$, a range that spans reasonable constant values in standardized space. For SR-SFC, SR-ALL, and SR-ALL-PC, the SR-ATM constants are held fixed. For all equations, the kernel parameters are held fixed at their NN-GAUSS values, so only the equation constants are refit. The optimized constants are rounded to four significant figures in standardized form, and these rounded constants define the evaluated models, so all results in the main text, including the physical constants in Table~S2, are computed from them.

\begin{table}[h!]
    \centering
    \small
    \begin{tabular}{cc}
    \toprule
    \textbf{Model} & \textbf{Constants} \\
    \midrule
    SR-BL & $\lambda = 1{,}687\,\mathrm{s^6/m^3}$, $B_c = -0.09673\,\mathrm{m/s^2}$, $\beta = 0.06791$ \\
    SR-ATM & $\lambda = 0.001137/\mathrm{K}^3$, $\kappa = 0.4141\,\mathrm{K/\%}$, $\gamma = 1.045$, $\Theta_c = -25.99\,\mathrm{K}$ \\
    SR-SFC & $\lambda_\mathrm{SHF} = 0.01251\,\mathrm{m^2/W}$, $\lambda_\mathrm{LHF} = 0.001645\,\mathrm{m^2/W}$, $\mathrm{LF}_c = 0.7598$ \\
    SR-ALL & $\lambda_{\theta_e} = 0.05863/\mathrm{K}$, $\lambda_\mathrm{SHF} = 0.07295\,\mathrm{m^2/W}$, $\mathrm{LF}_c = 0.7033$ \\
    SR-ALL-PC & $\lambda_{\theta_e} = 0.05863/\mathrm{K}$, $\lambda_\mathrm{SHF} = 0.07256\,\mathrm{m^2/W}$, $\mathrm{LF}_c = 0.7087$ \\
    \bottomrule
    \end{tabular}
    \vspace{1em}
    \caption{Physical constants of each symbolic regression model, derived from constants optimized in standardized form on the combined training and validation set (Text~S3). SR-SFC, SR-ALL, and SR-ALL-PC also contain the SR-ATM constants, and SR-ALL-PC is constructed from SR-ALL rather than discovered by PySR (Section~3.7). $\lambda$ in SR-BL, $\kappa$, $\lambda_\mathrm{SHF}$ in SR-SFC, and $\lambda_{\theta_e}$ are fixed by the standardization rather than fitted, so $\lambda_{\theta_e}$ is identical in SR-ALL and SR-ALL-PC. Constants without units are dimensionless.}
\end{table}

\noindent\textbf{Text S4. Physical Constraint Evaluation}

This text details the physical constraint evaluation summarized in Section~3.4 and Table~1. It describes how the constraints are tested in ERA5, why PC\textsubscript{1} is evaluated over land, and how the constraints are evaluated for the discovered equations. All evaluations use the test set (2018--2020).

\textit{Hypercube Binning in ERA5.} We use the hypercube binning approach of \citeA{Grundner2024} to test the constraints empirically. For each constraint, precipitation is related to one kernel-integrated feature (the tested feature) while the other two (the conditioning features) are held approximately constant (e.g., for PC\textsubscript{1}, $\widehat{\mathrm{RH}}$ is tested and $\widehat{\theta_e}$ and $\widehat{\theta_e^*}$ are conditioned on). Each conditioning feature is divided into $n$ equal-width intervals between its 1\textsuperscript{st} and 99\textsuperscript{th} percentiles over all grid points, with outlying values assigned to the nearest edge interval. Each pairing of one interval from each conditioning feature defines a hypercube, giving $n^2$ hypercubes in total, and hypercubes with fewer than 50 samples are discarded. Within each remaining hypercube, we regress ERA5 precipitation on the tested feature. The constraint is satisfied in a hypercube when the fitted slope has the expected sign or is zero, and the satisfaction rate is the fraction of retained hypercubes in which it is satisfied, with each hypercube weighted equally. Because only the kernel-integrated features are used for conditioning, covariation with LHF, SHF, and LF may also affect the slopes. To partly account for LF, hypercubes are formed separately over land and ocean using the same intervals, and the rate over all grid points is the fraction of all land and ocean hypercubes in which the constraint is satisfied. We apply the same procedure to the SR-BL expectation that precipitation does not decrease with $B_L$, except that $B_L$ has no other feature to condition on, so $B_L$ itself is divided into $n^2$ intervals and the slope is fitted within each.

Finer intervals hold the conditioning features more nearly constant but leave fewer samples for estimating each slope. We therefore average satisfaction rates across $n \in \{3,4,5,6,7\}$. Table~S3 gives the rates for each $n$ over land (LF $\geq$ 0.5), ocean (LF $<$ 0.5), and all grid points. PC\textsubscript{2} and PC\textsubscript{3} are both satisfied in every hypercube over both land and ocean. PC\textsubscript{1} is satisfied in 98.6\% of land hypercubes on average, but in only 61.2\% of ocean hypercubes, lowering the rate over all grid points to 79.9\%. Precipitation does not decrease with $B_L$ in 84.4\% of land intervals and 92.3\% of ocean intervals on average, and in 88.3\% over all grid points. We therefore use all grid points as the ERA5 reference for PC\textsubscript{2}, PC\textsubscript{3}, and $B_L$, but land alone as the reference for PC\textsubscript{1} (Table~1).

\begin{table}[h!]
    \centering
    \small
    \begin{tabular}{cccccccc}
    \toprule
    \textbf{Constraint} & \textbf{Region} & $n=3$ & $n=4$ & $n=5$ & $n=6$ & $n=7$ & \textbf{Average} \\
    \midrule
    \multirow{3}{*}{$\partial\mathrm{P}/\partial\widehat{\mathrm{RH}} \geq 0$}
        & Land & 100.0\% & 100.0\% & 100.0\% & 97.1\% & 95.7\% & \textbf{98.6\%} \\
        & Ocean & 66.7\% & 68.8\% & 58.3\% & 57.1\% & 55.3\% & 61.2\% \\
        & All & 83.3\% & 84.4\% & 79.2\% & 77.1\% & 75.5\% & 79.9\% \\
    \midrule
    \multirow{3}{*}{$\partial\mathrm{P}/\partial\widehat{\theta_e} \geq 0$}
        & Land & 100.0\% & 100.0\% & 100.0\% & 100.0\% & 100.0\% & 100.0\% \\
        & Ocean & 100.0\% & 100.0\% & 100.0\% & 100.0\% & 100.0\% & 100.0\% \\
        & All & 100.0\% & 100.0\% & 100.0\% & 100.0\% & 100.0\% & \textbf{100.0\%} \\
    \midrule
    \multirow{3}{*}{$\partial\mathrm{P}/\partial\widehat{\theta_e^*} \leq 0$}
        & Land & 100.0\% & 100.0\% & 100.0\% & 100.0\% & 100.0\% & 100.0\% \\
        & Ocean & 100.0\% & 100.0\% & 100.0\% & 100.0\% & 100.0\% & 100.0\% \\
        & All & 100.0\% & 100.0\% & 100.0\% & 100.0\% & 100.0\% & \textbf{100.0\%} \\
    \midrule
    \multirow{3}{*}{$\partial\mathrm{P}/\partial B_L \geq 0$}
        & Land & 100.0\% & 87.5\% & 84.0\% & 83.3\% & 67.3\% & 84.4\% \\
        & Ocean & 100.0\% & 100.0\% & 96.0\% & 91.7\% & 73.5\% & 92.3\% \\
        & All & 100.0\% & 93.8\% & 90.0\% & 87.5\% & 70.4\% & 88.3\% \\
    \bottomrule
    \end{tabular}
    \vspace{1em}
    \caption{ERA5 physical constraint satisfaction rates on the test set for each number of intervals per conditioning feature ($n$) and their average (Text~S4). Land and ocean correspond to LF $\geq$ 0.5 and LF $<$ 0.5, respectively, and all grid points combine the land and ocean hypercubes. Each rate is computed over at most $n^2$ hypercubes per region, so rates at small $n$ are coarse. Bold values are reported in Table~1.}
\end{table}

\textit{Ocean Violations of PC\textsubscript{1}.} The low ocean satisfaction rate for PC\textsubscript{1} does not indicate that oceanic precipitation decreases systematically with increasing $\widehat{\mathrm{RH}}$. Figure~S2 shows the slope $\mathrm{d}\mathrm{P}/\mathrm{d}\widehat{\mathrm{RH}}$ of every ocean hypercube, pooled across all values of $n$, against the fraction of samples in that hypercube with zero precipitation. Slopes are largest in hypercubes where precipitation is rarely zero and decrease toward zero as this dry fraction increases. All 54 violating hypercubes have zero precipitation in roughly 35\% or more of their samples, and their slopes are near zero. Satisfying hypercubes at similarly high dry fractions also have near-zero slopes, so in these mostly non-precipitating regimes the sign of the fitted slope is set by small fluctuations rather than by a physical response. Because the ocean violations are confined to these weak-response hypercubes, the ocean satisfaction rate does not provide a meaningful reference for PC\textsubscript{1}. Restricting PC\textsubscript{1} to land also does not favor the discovered equations, because each satisfies PC\textsubscript{1} for all input values, including at every ocean sample.

\textit{Derivatives of the Discovered Equations.} For any input feature $x$, differentiating Equation~2 gives:
\begin{equation*}
    \partial\mathrm{P}/\partial x =
    \begin{cases}
    s_y\exp(s_yz)(\partial z/\partial x), & z > 0, \\
    0, & z < 0.
    \end{cases}
\end{equation*}
Where $z < 0$, the derivative is zero and satisfies every constraint regardless of how the model responds to $x$, so counting these samples would inflate the satisfaction rates. We therefore evaluate the derivative of $\exp(s_yz)-1$ at every sample. Because $s_y\exp(s_yz)$ is always positive, this derivative has the same sign as $\partial z/\partial x$, which is the same whether $x$ is expressed in standardized or physical units because each predictor is standardized by dividing by a positive standard deviation. For SR-ATM and SR-SFC, this sign cannot violate any constraint because $\lambda$, $\kappa$, and $\gamma$ are positive (Section~3.5), so all three constraints hold for all input values. The derivatives of SR-ALL and SR-ALL-PC are examined in Section~3.7.

\begin{figure}[h!]
    \centering
    \includegraphics[width=2.64in]{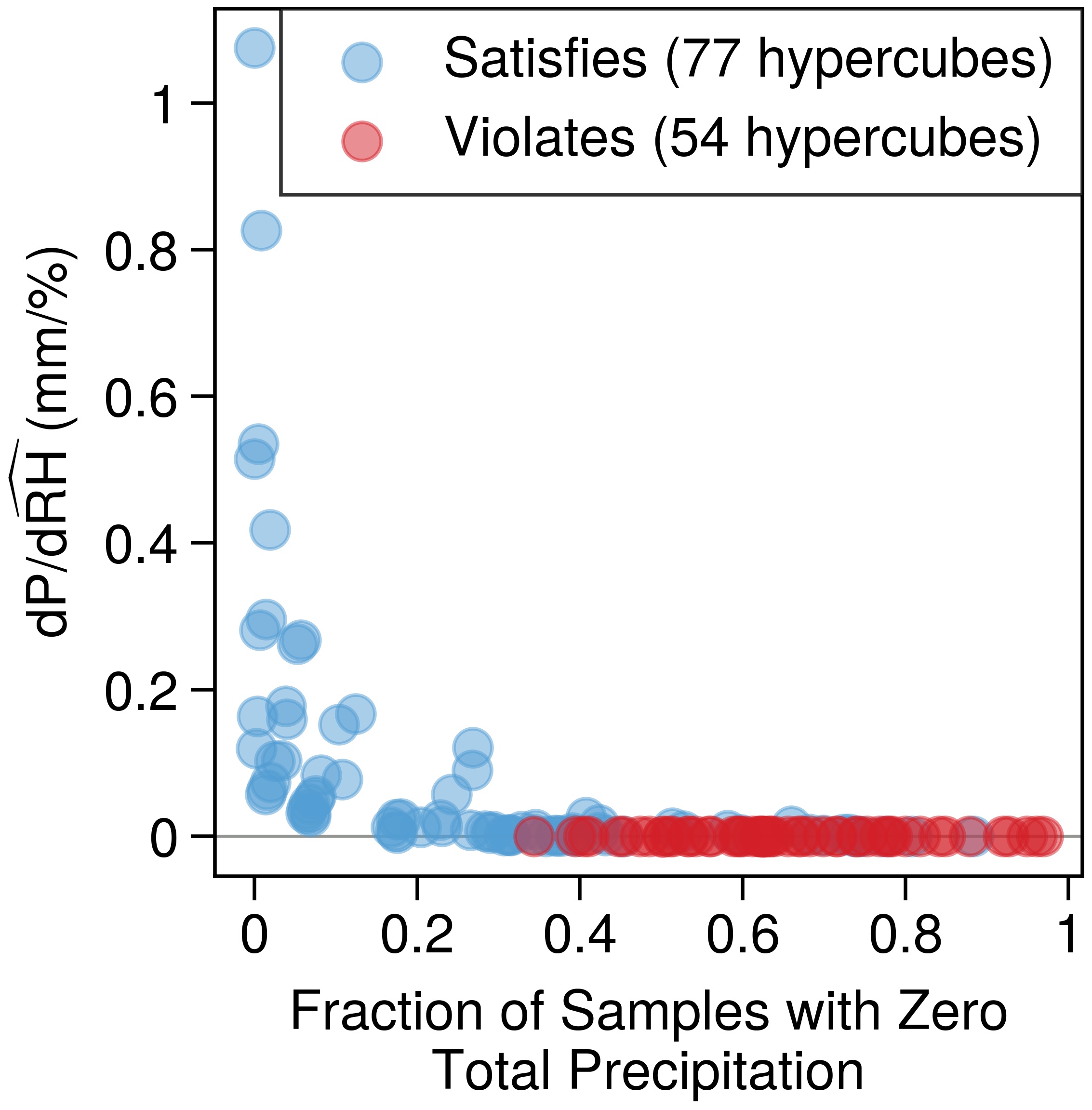}
    \caption{Slope of ERA5 precipitation with respect to $\widehat{\mathrm{RH}}$ in ocean hypercubes ($\mathrm{LF} < 0.5$) on the test set, pooled across all $n$ (Text~S4). Each marker is one hypercube, within which $\widehat{\theta_e}$ and $\widehat{\theta_e^*}$ are held approximately constant. The horizontal axis shows the fraction of samples in the hypercube with zero precipitation, and the vertical axis shows the slope of a linear fit of precipitation against $\widehat{\mathrm{RH}}$ within the hypercube. Blue markers have non-negative slopes and satisfy PC\textsubscript{1}, and red markers have negative slopes and violate it.}
\end{figure}

\end{document}